\documentclass{article} 
\usepackage{iclr2027_conference,times}

\usepackage{amsmath,amsfonts,bm}

\def\eqref#1{equation~\ref{#1}}

\def\1{\bm{1}}

\DeclareMathAlphabet{\mathsfit}{\encodingdefault}{\sfdefault}{m}{sl}
\SetMathAlphabet{\mathsfit}{bold}{\encodingdefault}{\sfdefault}{bx}{n}

\usepackage{algorithm}
\usepackage{algpseudocode}
\usepackage{hyperref}
\usepackage{url}
\usepackage{graphicx}
\usepackage{multirow}
\usepackage{booktabs}
\usepackage[table]{xcolor}
\usepackage[most]{tcolorbox}
\usepackage{amsthm}
\usepackage{wrapfig}
\definecolor{oursblue}{RGB}{225,242,249}
\definecolor{oursbg}{RGB}{249,247,237}

\title{Recurrent Self-Improvement: \\ Dynamic Cross-Loop On-Policy Distillation for Looped Language Models}

\author{
Yi Wang\textsuperscript{1} \quad
Rui Qian\textsuperscript{2} \quad
Yu Li\textsuperscript{3} \quad
Haoyang Yao\textsuperscript{4} \quad
Wenjie Wang\textsuperscript{1}
\\[0.4em]
\textsuperscript{1}ShanghaiTech University \quad
\textsuperscript{2}Fudan University
\\
\textsuperscript{3}Southeast University \quad
\textsuperscript{4}Peking University
}
\newcommand{\partitle}[1]{\smallskip\noindent \textbf{#1.}}
\newtheorem{proposition}{Proposition}
\newtheorem{theorem}{Theorem}
\iclrfinalcopy
\begin{document}
\maketitle
\lhead{Preprint}

\begin{abstract}
Looped Language Models (LoopLMs) offer a parameter efficient approach to scaling reasoning by reusing shared parameters across recurrent computation steps. Despite their promise, effective post-training of LoopLMs remains challenging.
Existing approaches either provide reward based supervision that is sparse or costly to extend across loops, or rely on external teachers or privileged information, leading to limited teacher availability or teacher-student context mismatch.
To address these limitations, we introduce \textbf{LoopOPD}, a cross-loop on-policy distillation framework that uses additional recurrent computation within a LoopLM as its own source of supervision. LoopOPD uses a frozen terminal loop policy as a compute privileged teacher for an intermediate loop student on student generated rollouts, providing dense supervision without an external teacher or privileged information.
We further propose Dynamic LoopOPD (\textbf{D-LoopOPD}), which continually refreshes the terminal loop teacher as the shared model parameters are updated, enabling recurrent self-improvement. We characterize how distillation updates propagate across loop depths and derive sufficient conditions under which a single update yields simultaneous local improvement at both loop depths.
Experiments on Ouro-Thinking models show that LoopOPD improves mathematical reasoning, while D-LoopOPD yields further gains through dynamic teacher updates. Despite being trained only on mathematical data, the resulting models also improve on general reasoning and code generation benchmarks, demonstrating that recurrent computation can serve as an effective source of supervision for LoopLMs. Our code and model checkpoints will be released upon acceptance.
\end{abstract}

\section{Introduction}
Looped Language Models (LoopLMs) have recently received growing attention as a promising approach to scaling reasoning through recurrent depth~\citep{geiping2025scaling}. They reuse shared Transformer blocks across multiple latent computation steps before emitting each token~\citep{zhu2025scaling}. This design increases computational depth without increasing parameter count, leading to stronger reasoning performance~\citep{saunshi2025latent}. However, effective post-training of LoopLMs remains challenging. Existing work applies or extends conventional post-training methods to LoopLMs, including reward based optimization and distillation, but faces limitations~\citep{williams2026prioritize}.

\begin{figure}[t]
    \centering
    \includegraphics[width=\linewidth]{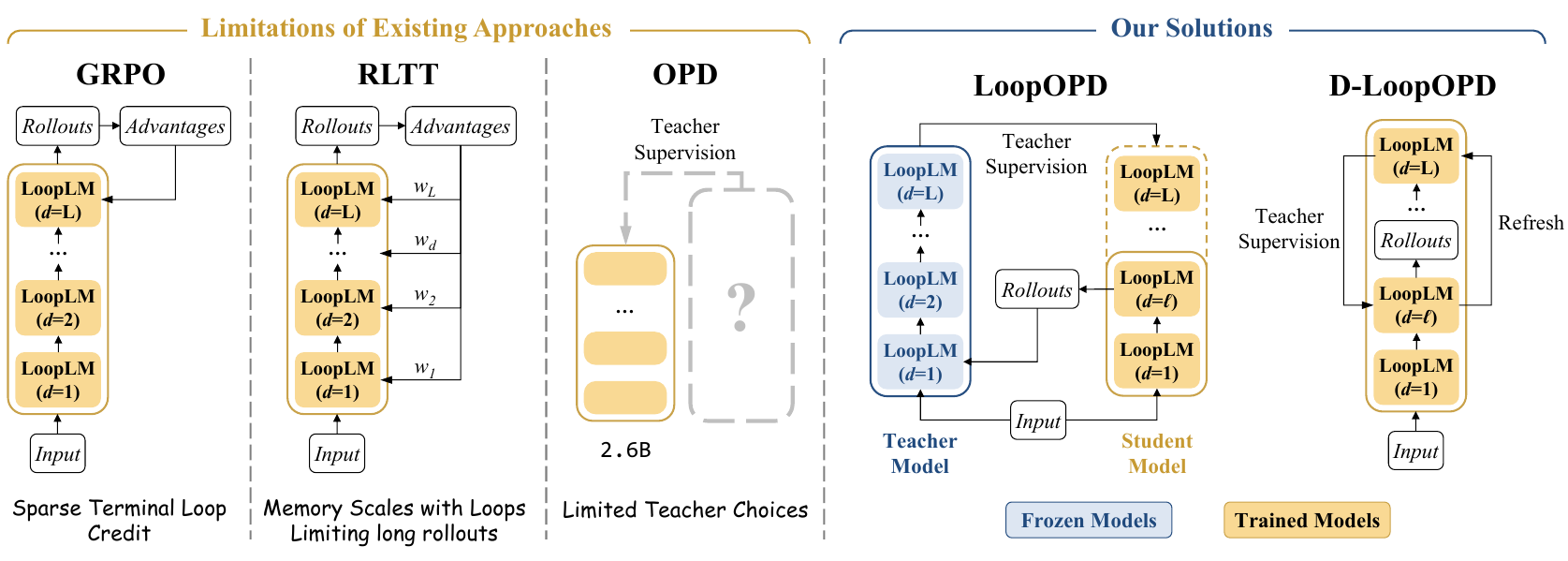}
    \caption{\textbf{Existing LoopLM post-training approaches vs. LoopOPD and D-LoopOPD.}
    Existing approaches variously face sparse supervision, increasing memory cost across loops, or dependence on stronger external teachers. LoopOPD uses additional recurrent computation as dense cross-loop supervision, while D-LoopOPD dynamically refreshes the terminal loop teacher during training.}
    \label{fig:intro}
    \vspace{-1em}
\end{figure}

Specifically, prior work on Ouro, a family of pretrained LoopLMs~\citep{zhu2025scaling}, reports limited gains from reinforcement learning with verifiable rewards (RLVR)~\citep{guo2025deepseek} over the supervised fine-tuning (SFT) checkpoint.
Group Relative Policy Optimization (GRPO)~\citep{shao2024deepseekmath} assigns reward credit only to the terminal loop, leaving intermediate latent computation without direct reward supervision. Reward Latent Thought Trajectories (RLTT)~\citep{williams2026prioritize} alleviates this limitation by extending reward credit across all loops, but requires retaining per loop log-probabilities, causing memory usage to scale linearly with the number of loops and thereby limiting long rollout training. On-Policy Distillation (OPD)~\citep{agarwal2024policy} provides dense distributional supervision but relies on an external, stronger teacher. For Ouro-2.6B-Thinking, no stronger same family LoopLM teacher is available, limiting OPD. On-Policy Self-Distillation (OPSD) instead constructs a same model teacher using privileged information~\citep{zhao2026self}, but introduces teacher--student context mismatch~\citep{kaur2026rethinking}. Figure~\ref{fig:intro} (left) summarizes the three main limitations motivating our work: sparse terminal loop credit, increasing memory cost across loops, and limited availability of stronger teachers.

Effective post-training of LoopLMs therefore calls for dense supervision that directly reaches intermediate loops when no stronger external teacher is available. Because deeper loops receive more recurrent computation, the terminal loop naturally provides a stronger policy within the same model. This raises a question: \textit{Can a LoopLM leverage its own additional recurrent computation to provide such supervision to its intermediate loops?}

To answer this question, we introduce LoopOPD. As illustrated in Figure~\ref{fig:intro} (right), LoopOPD uses an intermediate loop policy as the student and a frozen terminal loop policy as the compute privileged teacher. For student generated rollouts, both policies produce next token distributions on the same prefixes, and the terminal loop distribution is distilled into the intermediate loop through OPD. Unlike OPD and OPSD, the teacher's advantage arises not from additional parameters~\citep{li2026rethinking} or privileged information~\citep{zhao2026self}, but from additional recurrent computation. LoopOPD therefore turns the model's own recurrent computation into dense cross-loop supervision.

Intuitively, because different loops share the same parameters, training the student also updates the student model's terminal loop policy, creating a natural feedback path from the updated terminal policy back to the student. To exploit this feedback and further strengthen cross-loop supervision, we propose Dynamic LoopOPD (D-LoopOPD), which refreshes the terminal loop teacher at each optimization step, as shown in Figure~\ref{fig:intro} (right).
The updated terminal policy then supervises the next student update, forming a closed teacher-student feedback loop for recurrent self-improvement. We further analyze cross-loop update transport and derive sufficient conditions for a single update to induce simultaneous local descent of task surrogates at both the intermediate and terminal loops. Table~\ref{tab:supervision-comparison} compares the supervision signals and teacher requirements of these methods.

We evaluate LoopOPD and D-LoopOPD on Ouro-Thinking~\citep{zhu2025scaling} across mathematical reasoning~\citep{aime26}, general reasoning~\citep{hendrycks2021measuring}, and code generation benchmarks~\citep{liu2023evalplus}. On mathematical reasoning, both methods outperform existing baselines on average, with D-LoopOPD further improving over LoopOPD. Despite training only on mathematical data, they also improve general reasoning and code generation over the vanilla models.

Our contributions are threefold:
{
\setlength{\leftmargini}{3em}
\begin{itemize}
    \item We introduce LoopOPD, which uses a frozen terminal loop policy as a compute privileged teacher for an intermediate loop student, providing dense on-policy supervision without an external teacher or privileged information.

    \item We propose D-LoopOPD, which refreshes the terminal loop teacher at each optimization step through shared parameter updates, forming recurrent self-improvement. We further analyze cross-loop update transport and derive sufficient conditions for local improvement of task surrogates at both the intermediate and terminal loops.

    \item Experiments on Ouro-Thinking models show improved mathematical reasoning at both model scales. On Ouro-2.6B-Thinking, D-LoopOPD raises terminal loop AIME 2024 Avg@16 from 62.08 to 71.67, with cross-loop evaluation showing improvements across all loop depths. The resulting models also improve general reasoning and code generation despite training only on mathematical data.
\end{itemize}
}
\section{Related Work}
\label{sec:related-work}

\partitle{Looped and Recurrent-Depth Language Models}
Recurrent Transformer architectures such as the Universal Transformer~\citep{dehghani2018universal} explore parameter sharing across computational depth. Recent work studies this principle in language models, showing that repeated latent computation can improve reasoning and provide an additional axis for scaling test-time compute~\citep{saunshi2025latent, geiping2025scaling}. Ouro~\citep{zhu2025scaling} extends this line of work to pretrained LoopLMs, which recurrently apply shared Transformer blocks before token generation. Our work builds on this structure by using additional loop computation as a source of supervision.

\partitle{Intermediate Supervision for Recurrent Computation}
Recent work develops training objectives for recurrent latent computation. LSRL~\citep{ren-2025-lsrl} decodes intermediate recurrent states and scores them with an external grader to provide process supervision. RLTT~\citep{williams2026prioritize} extends outcome reward credit from the terminal loop to intermediate loop distributions in LoopLMs. LoopRPT~\citep{tang2026looprpt} uses an EMA teacher and noisy latent rollouts to construct stepwise reinforcement signals during pre-training, shaping intermediate representations. These methods highlight the value of directly supervising intermediate recurrent computation. Our work provides dense token level supervision to an intermediate loop using the model's own terminal loop distribution.

\partitle{On-Policy Distillation and Self-Distillation}
OPD trains a student on its own trajectories using dense teacher distributions, reducing the distribution mismatch from off-policy distillation~\citep{agarwal2024policy}. Recent analyses identify teacher-student compatibility and a teacher capability advantage as key factors for effective OPD~\citep{li2026rethinking}. OPSD constructs a teacher from the same model using privileged information~\citep{zhao2026self, hubotter2026reinforcement}, but introduce a context mismatch that can distort supervision~\citep{kaur2026rethinking}.
In contrast, LoopOPD retains the on-policy setting and teacher-student compatibility while deriving teacher advantage from additional recurrent computation, with teacher and student conditioned on the same context. D-LoopOPD further exploits shared parameters by refreshing the terminal loop teacher as the model is updated.

\begin{table}[t]
\caption{Comparison of supervision and teacher requirements across different methods.}
\vspace{-1em}
\label{tab:supervision-comparison}
\begin{center}
\small
\resizebox{\linewidth}{!}{
\begin{tabular}{lccccc}
\toprule
\raisebox{0.9ex}{\textbf{Method}}
& \shortstack{\textbf{Supervision}\\\textbf{Signal}}
& \shortstack{\textbf{No Verifiable}\\\textbf{Reward}}
& \shortstack{\textbf{No External}\\\textbf{Teacher}}
& \raisebox{-0.2ex}{\shortstack{\textbf{No Privileged}\\\textbf{Information}}}
& \raisebox{-0.2ex}{\shortstack{\textbf{Evolving}\\\textbf{Teacher}}} \\
\midrule
GRPO~\citep{shao2024deepseekmath} & Sparse & $\times$ & $\checkmark$ & $\checkmark$ & -- \\
RLTT~\citep{williams2026prioritize} & Relatively Dense & $\times$ & $\checkmark$ & $\checkmark$ & -- \\
OPD~\citep{agarwal2024policy} & Dense & $\checkmark$ & $\times$ & $\checkmark$ & $\times$ \\
OPSD~\citep{zhao2026self} & Dense & $\checkmark$ & $\checkmark$ & $\times$ & $\times$ \\
\rowcolor{oursbg}
LoopOPD & Dense & $\checkmark$ & $\checkmark$ & $\checkmark$ & $\times$ \\
\rowcolor{oursbg}
D-LoopOPD & Dense & $\checkmark$ & $\checkmark$ & $\checkmark$ & $\checkmark$ \\
\bottomrule
\end{tabular}
}
\end{center}
\vspace{-1em}
\end{table}
\section{Preliminaries}
\label{sec:preliminaries}
\partitle{Looped Language Models}
Let $x$ be a prompt and $y=(y_1,\ldots,y_m)$ a generated response of length $m$. A LoopLM recurrently applies shared weight Transformer blocks~\citep{vaswani2017attention, zhu2025scaling} for $L$ iterations before emitting each token. For the prediction of $y_i$, let $h_i^{(d)}$ denote the hidden state after loop $d$. Let $\theta$ denote the parameters of the LoopLM and $g_\theta$ its language model head. Applying $g_\theta$ to the loop $d$ hidden state defines the corresponding next token distribution
\begin{equation}
    P_\theta^{(d)}(\cdot\mid x,y_{<i})
    =
    \operatorname{softmax}\!\left(
        g_\theta(h_i^{(d)})
    \right),
    \qquad d=1,\ldots,L.
    \label{eq:loop}
\end{equation}
Standard autoregressive generation uses the terminal loop distribution, inducing the sequence policy
\begin{equation}
    \pi_\theta(y\mid x)
    =
    \prod_{i=1}^{m}
    P_\theta^{(L)}(y_i\mid x,y_{<i}).
    \label{eq:policy}
\end{equation}
For $d<L$, we refer to $P_\theta^{(d)}$ as the intermediate loop distributions.

\partitle{Group Relative Policy Optimization (GRPO)}
For each prompt $x$, GRPO samples a group of responses $\{y^{(j)}\}_{j=1}^{G}$ from the old policy
$\pi_{\theta_{\mathrm{old}}}$ and evaluates them with scalar rewards $r_1,\ldots,r_G$. The group relative advantage is $\widehat A_j=(r_j-\bar r)/\sigma_r$. For a LoopLM, response generation uses the terminal loop distribution, so the token level importance ratio is
\begin{equation}
    \rho_{j,i}(\theta)
    =
    \frac{
        P_\theta^{(L)}
        (y_i^{(j)}\mid x,y_{<i}^{(j)})
    }{
        P_{\theta_{\mathrm{old}}}^{(L)}
        (y_i^{(j)}\mid x,y_{<i}^{(j)})
    }.
    \label{eq:grpo-ratio}
\end{equation}
GRPO optimizes the clipped objective
{\small
\begin{equation}
\begin{aligned}
\mathcal{L}_{\mathrm{GRPO}}(\theta) ={}& -\mathbb{E}\!\left[\frac{1}{G}\sum_{j=1}^{G}\frac{1}{m_j}\sum_{i=1}^{m_j}\min\!\left(\rho_{j,i}(\theta)\widehat A_j,\operatorname{clip}(\rho_{j,i}(\theta),1-\epsilon,1+\epsilon)\widehat A_j\right)\right]+\beta D_{\mathrm{KL}}(\pi_\theta\|\pi_{\mathrm{ref}}),
\end{aligned}
\label{eq:grpo}
\end{equation}
}where $\epsilon$ is the clipping parameter, $\beta$ is the KL coefficient, and $\pi_{\mathrm{ref}}$ is the reference policy. Thus, GRPO uses the terminal loop policy for rollouts and direct reward credit, while gradients propagate through all loops to update the shared parameters.

\partitle{Reward Latent Thought Trajectories (RLTT)}
RLTT extends reward credit from the terminal loop to all loops while retaining the group relative advantage~\citep{williams2026prioritize}. Its policy objective is
{\small
\begin{equation}
\mathcal{L}_{\mathrm{RLTT}}(\theta)
=
-\mathbb{E}\!\left[
\frac{1}{G}\sum_{j=1}^{G}\frac{1}{m_j}\sum_{i=1}^{m_j}
\sum_{d=1}^{L}
\omega_d
\log P_\theta^{(d)}
(y_i^{(j)}\mid x,y_{<i}^{(j)})
\widehat A_j
\right]
+\beta D_{\mathrm{KL}}(\pi_\theta\|\pi_{\mathrm{ref}}),
\label{eq:rltt}
\end{equation}
}where $\omega_d\geq0$ and $\sum_{d=1}^{L}\omega_d=1$. A limitation of RLTT is that it retain per loop log-probabilities, leading to memory usage that scales linearly with the number of loops.

\partitle{On-Policy Distillation (OPD)}
OPD uses token level teacher distributions rather than scalar outcome rewards~\citep{agarwal2024policy}. Let $\theta$ and $\phi$ denote the parameters of the student LoopLM and the frozen teacher LoopLM. For $y\sim\pi_\theta(\cdot\mid x)$, both models are evaluated on the same prefixes $c_i=(x,y_{<i})$. The reverse-KL distillation objective is
{\small
\begin{equation}
\mathcal{L}_{\mathrm{OPD}}(\theta;\phi)
=
\mathbb{E}\!\left[
\frac{1}{m}\sum_{i=1}^{m}
D_{\mathrm{KL}}\!\left(
P_\theta^{(L)}(\cdot\mid c_i)
\,\middle\|\,
P_\phi^{(L)}(\cdot\mid c_i)
\right)
\right].
\label{eq:opd}
\end{equation}
}
\section{Method}
\label{sec:method}
Prior work reports limited gains from conventional reinforcement learning post-training on LoopLMs~\citep{zhu2025scaling,williams2026prioritize}. We explore an intrinsic source of supervision in LoopLMs: additional recurrent computation. Since a LoopLM admits a next token distribution at every loop, each loop naturally induces a corresponding autoregressive policy.

\subsection{LoopOPD: Cross-Loop On-Policy Distillation}
\label{sec:loopopd}
For any loop $d\leq L$, we define the loop $d$ policy as
\begin{equation}
    \pi_\theta^{(d)}(y\mid x)
    =
    \prod_{i=1}^{m}
    P_\theta^{(d)}(y_i\mid x,y_{<i}).
\end{equation}
As illustrated in Figure~\ref{fig:overview} (left), LoopOPD selects an intermediate loop $\ell<L$ as the student and uses the terminal loop $L$ policy as the compute privileged teacher.
We use $\ell=2$ and $L=4$.
Let $\theta_0$ denote the initial LoopLM parameters. We initialize the trainable student with $\theta\leftarrow\theta_0$ and the frozen teacher with $\bar{\theta}_0=\theta_0$. For each prompt $x$, a response is sampled from the student policy,
\begin{equation}
    y\sim\pi_\theta^{(\ell)}(\cdot\mid x).
\end{equation}
Both policies are evaluated on the same prefixes $c_i=(x,y_{<i})$, and LoopOPD distills the terminal loop distribution into the student using the reverse-KL objective in~\eqref{eq:opd}:
{\small
\begin{equation}
\mathcal{L}_{\mathrm{LoopOPD}}(\theta;\bar{\theta}_0)
=
\mathbb{E}\!\left[
\frac{1}{m}\sum_{i=1}^{m}
D_{\mathrm{KL}}\!\left(
P_\theta^{(\ell)}(\cdot\mid c_i)
\,\middle\|\,
P_{\bar{\theta}_0}^{(L)}(\cdot\mid c_i)
\right)
\right].
\end{equation}
}While LoopOPD supervises only the loop $\ell$ student policy, the loop $\ell$ and terminal loop policies of the trainable LoopLM share the same parameters. Updating the student therefore also changes the current terminal loop policy. Let $\theta_t$ and $\theta_{t+1}$ denote the parameters before and after optimization step $t$. This interaction can be summarized as
\begin{equation}
    \pi_{\bar{\theta}_0}^{(L)}
    \xrightarrow{\text{OPD}}
    \pi_{\theta_{t+1}}^{(\ell)}
    \xrightarrow{\text{parameter sharing}}
    \pi_{\theta_{t+1}}^{(L)}.
\end{equation}

\subsection{Dynamic LoopOPD: Recurrent Self-Improvement}
\label{sec:dynamic-loopopd}
\begin{figure}[t]
    \centering
    \includegraphics[width=\linewidth]{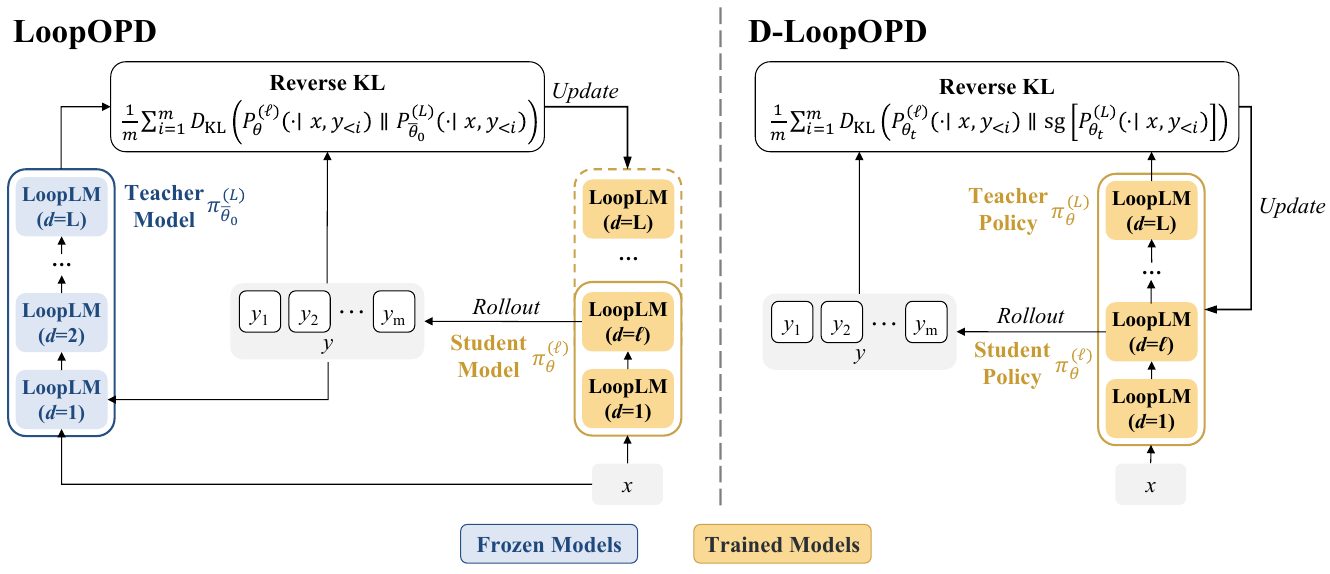}
    \vspace{-2em}
    \caption{\textbf{Overview of LoopOPD and D-LoopOPD.}
    LoopOPD uses a frozen terminal loop policy as a compute-privileged teacher and distills its next token distribution into an intermediate loop student on student generated rollouts. D-LoopOPD instead uses the current terminal loop policy of the same trainable LoopLM as the teacher, so each student update also changes the teacher used at the next step through shared parameters, forming recurrent self-improvement.}
    \label{fig:overview}
\end{figure}

As illustrated in Figure~\ref{fig:overview} (right), D-LoopOPD leverages this shared parameter coupling by using the current terminal loop policy as the teacher at each optimization step. Its objective is
{\small
\begin{equation}
\mathcal{L}_{\mathrm{D\text{-}LoopOPD}}(\theta)
=
\mathbb{E}\!\left[
\frac{1}{m}\sum_{i=1}^{m}
D_{\mathrm{KL}}\!\left(
P_\theta^{(\ell)}(\cdot\mid c_i)
\,\middle\|\,
\operatorname{sg}\!\left[P_\theta^{(L)}(\cdot\mid c_i)\right]
\right)
\right],
\label{eq:dynamic-loopopd}
\end{equation}
}where $\operatorname{sg}[\cdot]$ denotes stop-gradient. Gradients flow only through the student distribution, while the shared parameter update modifies both the loop $\ell$ and terminal loop policies. This interaction can be summarized as
\begin{equation}
    \pi_{\theta_t}^{(L)}
    \xrightarrow{\text{OPD}}
    \pi_{\theta_{t+1}}^{(\ell)}
    \xrightarrow{\text{parameter sharing}}
    \pi_{\theta_{t+1}}^{(L)}.
\end{equation}
The updated terminal loop policy then serves as the teacher for the next optimization step. Repeating this process forms a closed feedback loop between student learning and teacher evolution, which we refer to as recurrent self-improvement.

\subsection{Theoretical Analysis}
\label{sec:theoretical-analysis}
We analyze one distillation update and its effect at both loop depths. At step $t$, sample a batch $\mathcal{B}_t=\{(x^{(j)},y^{(j)})\}_{j=1}^{n_t}$ with $y^{(j)}\sim\pi_{\theta_t}^{(\ell)}(\cdot\mid x^{(j)})$, response lengths $m_j\geq1$, and prefixes $c_{j,i}=(x^{(j)},y_{<i}^{(j)})$.
We hold these sampled tokens and the targets $q_t(\cdot\mid c)=\operatorname{sg}[P_{\theta_t}^{(L)}(\cdot\mid c)]$ fixed during the update, as in OPD.
The empirical loss retains the response-wise normalization of \eqref{eq:dynamic-loopopd}:
{\small
\begin{equation}
\begin{aligned}
\mathcal{L}_t(\theta)&=
\frac{1}{n_t}\sum_{j=1}^{n_t}\frac{1}{m_j}\sum_{i=1}^{m_j}
D_{\mathrm{KL}}\!\left(
P_\theta^{(\ell)}(\cdot\mid c_{j,i})
\,\middle\|\,
q_t(\cdot\mid c_{j,i})
\right),\\
g_t&=\nabla_\theta\mathcal{L}_t(\theta_t),
\qquad
\theta_{t+1}=\theta_t-\eta_t g_t.
\end{aligned}
\label{eq:step-local-objective}
\end{equation}}where $g_t$ is the gradient of the empirical distillation loss at $\theta_t$, and $\eta_t>0$ is the step size. We analyze an exact gradient step for the full-distribution KL objective.

\begin{proposition}[Descent with a step-frozen teacher]
\label{prop:absorption}
Suppose $\mathcal{L}_t$ has an $M_t$-Lipschitz gradient, $M_t>0$, in a
neighborhood of the update segment. If $g_t\neq0$ and $0<\eta_t<2/M_t$, then
\begin{equation}
\mathcal{L}_t(\theta_{t+1})
\leq
\mathcal{L}_t(\theta_t)
-\eta_t\!\left(1-\frac{M_t\eta_t}{2}\right)\!\|g_t\|_2^2
<\mathcal{L}_t(\theta_t).
\label{eq:absorption-bound}
\end{equation}
\end{proposition}
This guarantees a decrease in the batch-averaged KL divergence to the step-frozen teacher $q_t$.

For $d\in\{\ell,L\}$, concatenate the logits $g_\theta(h_i^{(d)})$ on the batch prefixes into $z_d(\theta)$, and let $J_d=\partial z_d/\partial\theta|_{\theta_t}$ and $r_t=\nabla_{z_\ell}\mathcal{L}_t|_{\theta_t}$.
With the teacher stopped, $g_t=J_\ell^\top r_t$. If the logit maps have locally Lipschitz Jacobians, then
\begin{equation}
z_d(\theta_{t+1})-z_d(\theta_t)
=-\eta_t
\underbrace{J_dJ_\ell^\top}_{K_{d\ell}}
r_t
+O\!\left(\eta_t^2\|g_t\|_2^2\right),
\qquad d\in\{\ell,L\},
\label{eq:cross-loop-transport}
\end{equation}
where $K_{d\ell}$ is a block of the empirical tangent kernel~\citep{jacot2018neural}. 
Parameter sharing gives the first order terminal logit change $-\eta_tK_{L\ell}r_t$, whose effect on the task surrogate depends on gradient alignment.

For analysis, let $\mathcal{R}_d(\theta)$ be a differentiable task surrogate computed from the logits at loop depth $d$ on a fixed evaluation set $\mathcal{E}$, such as gold-token negative log-likelihood. Define
\begin{equation}
\gamma_{d,t}
=\left\langle\nabla_\theta\mathcal{R}_d(\theta_t),g_t\right\rangle,
\qquad d\in\{\ell,L\}.
\label{eq:risk-alignment}
\end{equation}
Here $\gamma_{d,t}$ measures the alignment between the distillation gradient and the task surrogate gradient at depth $d$; $\gamma_{d,t}>0$ means that $-g_t$ is a descent direction for $\mathcal{R}_d$.

\begin{theorem}[Joint local decrease of task surrogates]
\label{thm:joint-improvement}
Suppose $\mathcal{R}_d$ has a $\beta_d$-Lipschitz gradient, $\beta_d>0$, in a neighborhood of the update segment for each $d\in\{\ell,L\}$. If $\gamma_{\ell,t}>0$, $\gamma_{L,t}>0$, and
\begin{equation}
0<\eta_t<
\min_{d\in\{\ell,L\}}
\frac{2\gamma_{d,t}}{\beta_d\|g_t\|_2^2},
\label{eq:joint-step-size}
\end{equation}
then $\mathcal{R}_\ell(\theta_{t+1})<\mathcal{R}_\ell(\theta_t)$ and $\mathcal{R}_L(\theta_{t+1})<\mathcal{R}_L(\theta_t)$ simultaneously.
\end{theorem}
For evaluation logits $\widetilde z_d$ on $\mathcal{E}$, write $\widetilde J_d=\partial\widetilde z_d/\partial\theta|_{\theta_t}$ and $a_{d,t}=\nabla_{\widetilde z_d}\mathcal{R}_d|_{\theta_t}$. Then
\begin{equation}
\gamma_{d,t}
=
a_{d,t}^{\top}\widetilde J_dJ_\ell^\top r_t,
\qquad d\in\{\ell,L\}.
\label{eq:evaluation-transport}
\end{equation}
This expression links the training time distillation update to the evaluation task gradients at both loop depths. Appendix~\ref{app:theory-proofs} provides more details.
\section{Experiments}
\label{sec:experiments}
\subsection{Experimental Setup}
\label{sec:experimental-setup}
\partitle{Models}
We conduct experiments on Ouro-1.4B-Thinking and Ouro-2.6B-Thinking~\citep{zhu2025scaling}. For comparison across model scales, we include Qwen3-1.7B, Qwen3-4B, Qwen3-8B~\citep{qwen3}, Qwen3.5-4B, and Qwen3.5-9B~\citep{qwen35blog} in thinking mode.

\partitle{Baselines and Training Dataset}
For Ouro, we compare LoopOPD and D-LoopOPD against GRPO~\citep{shao2024deepseekmath}, RLTT~\citep{williams2026prioritize}, and OPSD~\citep{zhao2026self}. We evaluate OPD~\citep{agarwal2024policy} on Ouro-1.4B-Thinking, using Ouro-2.6B-Thinking as the teacher. For Qwen3, we evaluate the vanilla models and GRPO variants, while for Qwen3.5 we evaluate the vanilla models. All post-training methods are trained on DAPO-Math-17k~\citep{yu2025dapo}.

\partitle{Evaluation Benchmarks}
We evaluate mathematical reasoning on AIME 2024~\citep{aime24}, AIME 2025~\citep{aime25}, AIME 2026~\citep{aime26}, OlympiadBench~\citep{he2024olympiadbench}, and BeyondAIME~\citep{bytedance}. For general reasoning and code generation, we evaluate on MMLU~\citep{hendrycks2021measuring}, MBPP and MBPP+~\citep{austin2021program, liu2023evalplus}, and HumanEval and HumanEval+~\citep{chen2021codex, liu2023evalplus}. Detailed training and evaluation settings are provided in Appendix~\ref{app:details}.

\subsection{Main Results}
\label{sec:main-results}
\partitle{Mathematical Reasoning}
As shown in Table~\ref{tab:math-results}, LoopOPD and D-LoopOPD improve mathematical reasoning on Ouro-2.6B-Thinking. Averaging Avg@16 across AIME 2024, AIME 2025, and AIME 2026, the vanilla model and GRPO achieve 55.49 and 56.60, while LoopOPD and D-LoopOPD improve to 59.37 and 62.36. D-LoopOPD achieves the best Avg@16 on all three AIME benchmarks within the Ouro-2.6B group, improving over LoopOPD by 6.46, 1.46, and 1.05 points on AIME 2024, AIME 2025, and AIME 2026, respectively. On AIME 2024, D-LoopOPD reaches 71.67 Avg@16, surpassing vanilla Qwen3-8B at 71.25 and approaching vanilla Qwen3.5-9B at 72.08 with fewer parameters.
On OlympiadBench, D-LoopOPD also achieves the best result at 65.77, while its BeyondAIME score of 34.00 remains comparable to those of the vanilla model and GRPO.
Similar improvements are observed on Ouro-1.4B-Thinking, where LoopOPD and D-LoopOPD outperform the vanilla model, GRPO, and OPD on average across the three AIME benchmarks; LoopOPD achieves the best OlympiadBench score and D-LoopOPD the best BeyondAIME score.
\begin{table*}[h]
\caption{\textbf{Performance comparison on mathematical reasoning benchmarks.}
AIME24, AIME25, and AIME26 are reported with Avg@16 and Pass@16, while OlympiadBench and BeyondAIME are reported with accuracy. All evaluations use a maximum response length of 31,744 tokens. Best results within each Ouro model group are shown in \textbf{bold}.}
\vspace{-1em}
\label{tab:math-results}
\begin{center}
\resizebox{\textwidth}{!}{
\begin{tabular}{llcccccccc}
\toprule
\multirow{2}{*}{\raisebox{-0.7ex}{\textbf{Model}}}
& \multirow{2}{*}{\raisebox{-0.7ex}{\textbf{Method}}}
& \multicolumn{2}{c}{\textbf{AIME24}}
& \multicolumn{2}{c}{\textbf{AIME25}}
& \multicolumn{2}{c}{\textbf{AIME26}}
& \multirow{2}{*}{\raisebox{-3ex}{\shortstack{\textbf{Olympiad}\\\textbf{Bench}}}}
& \multirow{2}{*}{\raisebox{-3ex}{\shortstack{\textbf{Beyond}\\\textbf{AIME}}}} \\
\cmidrule(lr){3-4}\cmidrule(lr){5-6}\cmidrule(lr){7-8}
& & Avg@16 & Pass@16
& Avg@16 & Pass@16
& Avg@16 & Pass@16
& & \\
\midrule
\multirow{2}{*}{Qwen3-1.7B (Thinking)}
& Vanilla
& 47.92 & 76.67 & 36.46 & 56.67 & 38.13 & 76.67 & 56.30 & 15.00 \\
& GRPO
& 51.04 & 73.33 & 37.50 & 56.67 & 40.42 & 80.00 & 57.19 & 17.00 \\
\midrule
\multirow{2}{*}{Qwen3-4B (Thinking)}
& Vanilla
& 61.25 & 83.33 & 52.08 & 73.33 & 52.92 & 76.67 & 68.44 & 31.00 \\
& GRPO
& 65.42 & 83.33 & 51.67 & 76.67 & 55.21 & 80.00 & 68.89 & 32.00 \\
\midrule
Qwen3.5-4B (Thinking)\rule[-1.0ex]{0pt}{3.2ex}
& Vanilla
& 54.79 & 90.00 & 51.46 & 73.33 & 55.00 & 80.00 & 61.63 & 23.00 \\
\midrule
\multirow{7}{*}{
\shortstack[c]{
Ouro-1.4B-Thinking\\
($d=4$)
}}
& Vanilla
& 51.25 & 73.33 & 41.25 & 70.00 & 45.21 & 73.33 & 59.40 & 24.00 \\
& GRPO
& 51.88 & 76.67 & 40.21 & 70.00 & 46.46 & 70.00 & 60.14 & 24.00 \\
& RLTT
& 50.00 & 80.00 & 37.50 & 66.67 & 41.87 & 66.67 & 60.44 & 22.00 \\
& OPSD
& 49.58 & 73.33 & 41.04 & 63.33 & 44.79 & 70.00 & 60.59 & 22.00 \\
& OPD
& 50.42 & 83.33 & 38.54 & 66.67 & 46.25 & 73.33 & 60.44 & 25.00 \\
& \cellcolor{oursbg}LoopOPD
& \cellcolor{oursbg}53.33
& \cellcolor{oursbg}80.00
& \cellcolor{oursbg}\textbf{43.54}
& \cellcolor{oursbg}70.00
& \cellcolor{oursbg}46.25
& \cellcolor{oursbg}73.33
& \cellcolor{oursbg}\textbf{60.74}
& \cellcolor{oursbg}24.00 \\
& \cellcolor{oursbg}D-LoopOPD
& \cellcolor{oursbg}\textbf{55.00}
& \cellcolor{oursbg}\textbf{86.67}
& \cellcolor{oursbg}\textbf{43.54}
& \cellcolor{oursbg}\textbf{73.33}
& \cellcolor{oursbg}\textbf{47.92}
& \cellcolor{oursbg}\textbf{76.67}
& \cellcolor{oursbg}60.59
& \cellcolor{oursbg}\textbf{27.00} \\
\midrule
\multirow{2}{*}{Qwen3-8B (Thinking)}
& Vanilla
& 71.25 & 90.00 & 64.38 & 83.33 & 66.67 & 86.67 & 69.48 & 37.00 \\
& GRPO
& 73.54 & 93.33 & 68.96 & 83.33 & 69.38 & 86.67 & 71.26 & 38.00 \\
\midrule
Qwen3.5-9B (Thinking)\rule[-1.0ex]{0pt}{3.2ex}
& Vanilla
& 72.08 & 93.33 & 61.46 & 86.67 & 66.04 & 86.67 & 64.89 & 36.00 \\
\midrule
\multirow{6}{*}{
\shortstack[c]{
Ouro-2.6B-Thinking\\
($d=4$)
}}
& Vanilla
& 62.08 & 83.33 & 48.96 & 73.33 & 55.42 & 83.33 & 64.30 & \textbf{35.00} \\
& GRPO
& 62.92 & 86.67 & 49.79 & 73.33 & 57.08 & 86.67 & 64.59 & \textbf{35.00} \\
& RLTT
& 56.67 & 83.33 & 36.46 & 66.67 & 46.04 & 76.67 & 60.29 & 28.00 \\
& OPSD
& 60.42 & 83.33 & 43.13 & 70.00 & 50.62 & 73.33 & 63.26 & 30.00 \\
& \cellcolor{oursbg}LoopOPD
& \cellcolor{oursbg}65.21
& \cellcolor{oursbg}\textbf{90.00}
& \cellcolor{oursbg}53.75
& \cellcolor{oursbg}\textbf{80.00}
& \cellcolor{oursbg}59.17
& \cellcolor{oursbg}\textbf{90.00}
& \cellcolor{oursbg}64.88
& \cellcolor{oursbg}33.00 \\
& \cellcolor{oursbg}D-LoopOPD
& \cellcolor{oursbg}\textbf{71.67}
& \cellcolor{oursbg}\textbf{90.00}
& \cellcolor{oursbg}\textbf{55.21}
& \cellcolor{oursbg}\textbf{80.00}
& \cellcolor{oursbg}\textbf{60.21}
& \cellcolor{oursbg}86.67
& \cellcolor{oursbg}\textbf{65.78}
& \cellcolor{oursbg}34.00 \\
\bottomrule
\end{tabular}
}
\end{center}
\vspace{-1em}
\end{table*}

\partitle{General Reasoning and Code Generation}
Table~\ref{tab:nonmath-results} further examines performance beyond mathematical reasoning. On Ouro-2.6B-Thinking, D-LoopOPD improves over the vanilla model on all five benchmarks, with gains of 0.58 on MMLU, 1.06 on MBPP, 1.33 on MBPP+, 0.61 on HumanEval, and 0.61 on HumanEval+. It achieves the best results within the model group on MMLU, MBPP, and MBPP+. Similar gains are observed on Ouro-1.4B-Thinking, where LoopOPD achieves the best MMLU result, while D-LoopOPD achieves the best results on MBPP, MBPP+, and HumanEval+. These gains extend to general reasoning and code generation even though post-training uses only mathematical data.
\begin{table*}[t]
\caption{\textbf{Performance comparison on general reasoning and code generation benchmarks.}
MMLU is reported with accuracy, while MBPP, MBPP+, HumanEval, and HumanEval+ are reported with Pass@1. Best results within each Ouro model group are shown in \textbf{bold}.}
\label{tab:nonmath-results}
\begin{center}
\resizebox{0.85\textwidth}{!}{
\begin{tabular}{llccccc}
\toprule
\textbf{Model}\rule[-1.0ex]{0pt}{3.2ex}
& \textbf{Method}
& \textbf{MMLU}
& \textbf{MBPP}
& \textbf{MBPP+}
& \textbf{HumanEval}
& \textbf{HumanEval+} \\
\midrule
\multirow{2}{*}{Qwen3-1.7B (Thinking)}
& Vanilla
& 60.21 & 59.26 & 50.26 & 69.51 & 63.41 \\
& GRPO
& 61.00 & 60.05 & 50.00 & 71.34 & 64.63 \\
\midrule
\multirow{2}{*}{Qwen3-4B (Thinking)}
& Vanilla
& 70.07 & 68.78 & 59.79 & 77.44 & 70.73 \\
& GRPO
& 68.49 & 67.99 & 60.85 & 75.61 & 69.51 \\
\midrule
Qwen3.5-4B (Thinking)\rule[-1.0ex]{0pt}{3.2ex}
& Vanilla
& 69.66 & 75.40 & 61.11 & 82.93 & 76.83 \\
\midrule
\multirow{7}{*}{
\shortstack[c]{
Ouro-1.4B-Thinking\\
($d=4$)
}}
& Vanilla
& 67.78 & 72.22 & 62.70 & 74.39 & 70.12 \\
& GRPO
& 67.51 & 72.49 & 62.70 & 75.00 & \textbf{71.95} \\
& RLTT
& 68.01 & 73.02 & 62.70 & \textbf{77.44} & 71.34 \\
& OPSD
& 66.50 & 71.96 & 61.11 & 74.39 & 69.51 \\
& OPD
& 67.83 & 72.22 & 62.43 & 74.39 & 70.12\\
& \cellcolor{oursbg}LoopOPD
& \cellcolor{oursbg}\textbf{68.05}
& \cellcolor{oursbg}72.22
& \cellcolor{oursbg}62.70
& \cellcolor{oursbg}76.22
& \cellcolor{oursbg}71.34 \\
& \cellcolor{oursbg}D-LoopOPD
& \cellcolor{oursbg}67.87
& \cellcolor{oursbg}\textbf{73.28}
& \cellcolor{oursbg}\textbf{63.23}
& \cellcolor{oursbg}76.22
& \cellcolor{oursbg}\textbf{71.95} \\
\midrule
\multirow{2}{*}{Qwen3-8B (Thinking)}
& Vanilla
& 74.89 & 71.69 & 62.43 & 84.15 & 73.78 \\
& GRPO
& 74.48 & 71.96 & 63.23 & 85.37 & 75.00 \\
\midrule
Qwen3.5-9B (Thinking)\rule[-1.0ex]{0pt}{3.2ex}
& Vanilla
& 69.90 & 79.10 & 64.81 & 86.59 & 82.32 \\
\midrule
\multirow{6}{*}{
\shortstack[c]{
Ouro-2.6B-Thinking\\
($d=4$)
}}
& Vanilla
& 76.09 & 74.87 & 62.43 & 78.66 & 74.39 \\
& GRPO
& 76.32 & 75.13 & 62.17 & 78.05 & 74.39 \\
& RLTT
& 76.11 & 75.40 & 61.90 & \textbf{81.71} & \textbf{77.44} \\
& OPSD
& 75.84 & 74.87 & 62.70 & 76.83 & 72.56 \\
& \cellcolor{oursbg}LoopOPD
& \cellcolor{oursbg}76.09
& \cellcolor{oursbg}75.40
& \cellcolor{oursbg}62.70
& \cellcolor{oursbg}78.66
& \cellcolor{oursbg}74.39\\
& \cellcolor{oursbg}D-LoopOPD
& \cellcolor{oursbg}\textbf{76.67}
& \cellcolor{oursbg}\textbf{75.93}
& \cellcolor{oursbg}\textbf{63.76}
& \cellcolor{oursbg}79.27
& \cellcolor{oursbg}75.00 \\
\bottomrule
\end{tabular}
}
\end{center}
\end{table*}

\subsection{Training Dynamics and Cross-Loop Performance}
\label{sec:training-dynamics}
\begin{figure}[h]
    \centering
    \includegraphics[width=\linewidth]{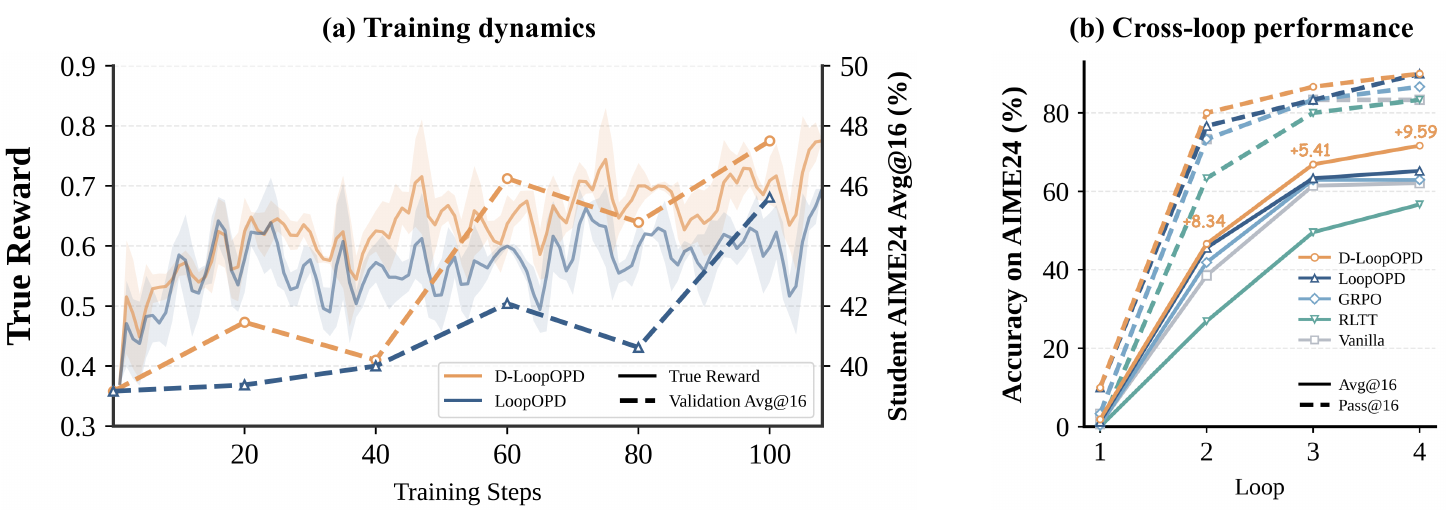}
    \vspace{-2em}
    \caption{\textbf{Training dynamics and cross-loop performance of Ouro-2.6B-Thinking.} (a) Training reward and student loop validation Avg@16 for LoopOPD and D-LoopOPD over training. (b) Avg@16 and Pass@16 across loop depths for different methods. All evaluations use a maximum response length of 31,744 tokens. Exact values are provided in Appendix~\ref{app:supp-experiments}.}
    \label{fig:training_dynamics}
\end{figure}

\partitle{Training Dynamics}
Figure~\ref{fig:training_dynamics} (a) compares LoopOPD and D-LoopOPD during training. Both the rollout reward and validation performance are measured at the student loop $\ell=2$. D-LoopOPD achieves higher training reward and higher validation Avg@16 than LoopOPD. These trends support the benefit of dynamically using the current terminal loop policy as the teacher for student learning.
\partitle{Cross-Loop Performance}
Figure~\ref{fig:training_dynamics} (b) evaluates the trained models across all four loops. Both LoopOPD and D-LoopOPD improve Avg@16 over the vanilla model at every depth, with D-LoopOPD achieving the highest Avg@16 among the compared methods. At the directly supervised loop 2, D-LoopOPD raises Avg@16 from 39.16 to 47.50, a gain of 8.34 percentage points. The gains extend to loops 3 and 4, where Avg@16 increases by 5.41 and 9.59 points to 66.87 and 71.67. These results support supervising intermediate computation as an effective way to improve both intermediate and terminal policies.

\begin{figure}[h]
    \centering
    \includegraphics[width=0.47\linewidth]{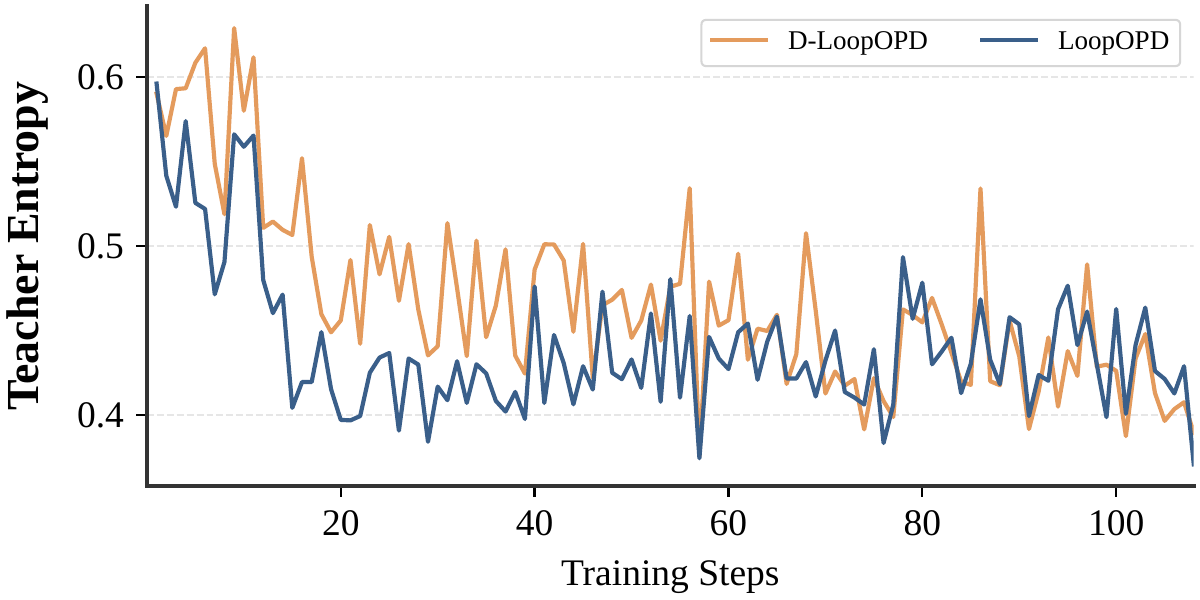}
    \hfill
    \includegraphics[width=0.47\linewidth]{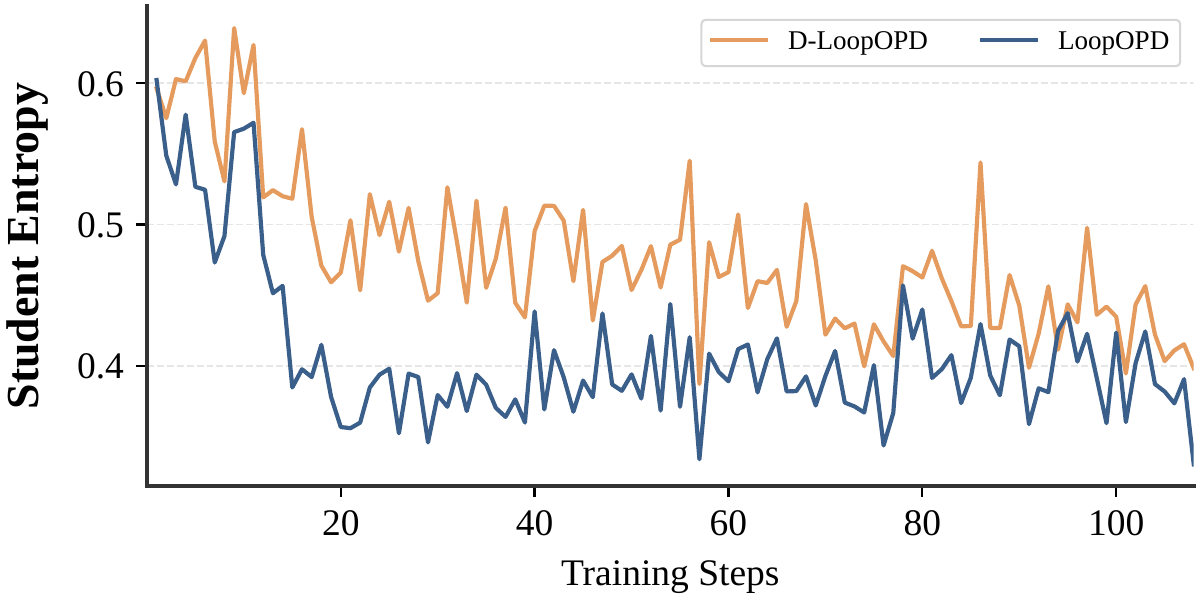}
    \vspace{-1em}
    \caption{Teacher and student entropy during training for Ouro-2.6B-Thinking. \textbf{Left:} entropy of the terminal loop teacher. \textbf{Right:} entropy of the student loop $\ell=2$. Both are measured on student generated rollouts.}
    \label{fig:entropy-dynamics}
\end{figure}

\partitle{Teacher and Student Entropy}
Figure~\ref{fig:entropy-dynamics} compares the teacher and student token distribution entropy of LoopOPD and D-LoopOPD during training.
D-LoopOPD maintains higher teacher and student entropy than LoopOPD through most of training, with the gap narrowing toward later steps. This suggests that dynamically refreshing the terminal loop teacher maintains less peaked distributions during training, which support more diverse student rollouts and broader exploration~\citep{jin-etal-2026-revisiting, cui2025entropy}. More details are provided in Appendix~\ref{app:supp-experiments}.

\subsection{Ablation Study}
\label{sec:ablation}
\begin{wrapfigure}{r}{0.5\textwidth}
    \vspace{-14pt}
    \centering
    \includegraphics[width=\linewidth]{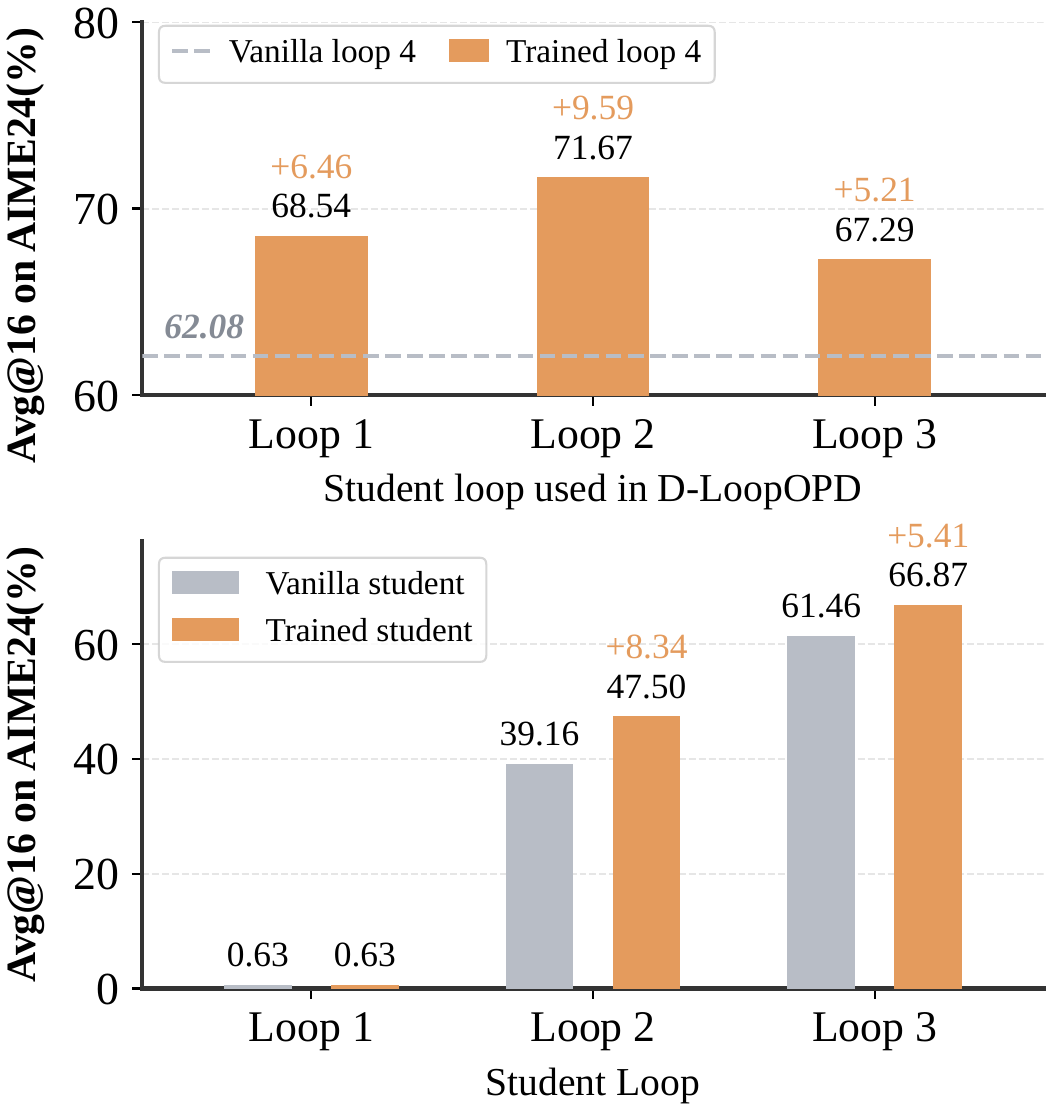}
    \vspace{-2em}
    \caption{Student depth ablation for D-LoopOPD on Ouro-2.6B-Thinking. AIME 2024 Avg@16 at the student and terminal loops.}
    \label{fig:student-depth-ablation}
    \vspace{-14pt}
\end{wrapfigure}
We study the effect of student depth in D-LoopOPD by using loop 1, loop 2, or loop 3 as the student while keeping the terminal teacher at loop 4. As shown in Figure~\ref{fig:student-depth-ablation}, loop 2 yields the best terminal performance, improving loop 4 Avg@16 from 62.08 to 71.67. Loop 3 already starts close to the initial terminal policy, with Avg@16 of 61.46 versus 62.08, leaving less room for distillation. In contrast, loop 1 has a much larger gap to the teacher but very low standalone accuracy. Its student Avg@16 remains at 0.63 after training, although the terminal loop improves to 68.54. Loop 2 provides a favorable balance between student capability and teacher advantage: its student Avg@16 improves from 39.16 to 47.50, while the terminal loop gains 9.59 points. These results indicate that neither a larger teacher-student gap nor a stronger student alone determines the effectiveness of cross-loop distillation. Overall, the intermediate loop $\ell=2$ yields the strongest terminal loop improvement in our setting.
\section{Conclusion}
We propose LoopOPD, which uses a frozen terminal loop policy to supervise an intermediate loop student through on-policy distillation, without an external teacher or privileged information. D-LoopOPD instead uses the current terminal loop policy as the teacher, allowing student updates to shape subsequent supervision through shared parameters. Our analysis characterizes this interaction and gives sufficient conditions for a single gradient update to improve task surrogates locally at both loop depths. Experiments on Ouro-Thinking models show gains in mathematical reasoning, with D-LoopOPD further improving mathematical performance over LoopOPD. Cross-loop evaluation shows that the gains extend beyond the directly supervised student to deeper loops. D-LoopOPD also improves general reasoning and code generation over the vanilla models despite using only mathematical training data. These findings support additional recurrent computation as an effective source of supervision for post-training LoopLMs.

\bibliography{iclr2027_conference}
\bibliographystyle{iclr2027_conference}

\clearpage
\appendix
\section{Proofs}
\label{app:theory-proofs}
\partitle{Scope and Normalization}
We condition on the rollout batch $\mathcal{B}_t$ in \eqref{eq:step-local-objective}. The sampled tokens, response lengths, and teacher probabilities are held fixed during differentiation. Hidden states and logits on these fixed token prefixes are functions of the trainable parameters. We use the reverse KL over the softmax distributions in \eqref{eq:loop}, assuming finite logits and a finite vocabulary so that all probabilities are positive. Each token occurrence $(j,i)$ has weight $w_{j,i}=1/(n_tm_j)$, so that $\sum_j\sum_i w_{j,i}=1$. These weights match the per response normalization in \eqref{eq:dynamic-loopopd}.

\partitle{Reverse-KL Gradient}
For a token occurrence $(j,i)$, abbreviate
\begin{equation}
\begin{aligned}
p_{j,i}(v)&=P_{\theta_t}^{(\ell)}(v\mid c_{j,i}),
&q_{j,i}(v)&=q_t(v\mid c_{j,i}),\\
\kappa_{j,i}&=\sum_{v\in\mathcal{V}}p_{j,i}(v)
\log\frac{p_{j,i}(v)}{q_{j,i}(v)}.
\end{aligned}
\label{eq:appendix-token-kl}
\end{equation}
Let $z_{\ell,j,i,v}$ be the corresponding student logit. The block of
$r_t=\nabla_{z_\ell}\mathcal{L}_t|_{\theta_t}$ is
\begin{equation}
(r_t)_{j,i,v}=w_{j,i}p_{j,i}(v)
\left(\log\frac{p_{j,i}(v)}{q_{j,i}(v)}-\kappa_{j,i}\right).
\label{eq:reverse-kl-logit-gradient}
\end{equation}
To verify this, use
$\partial p(u)/\partial z_v=p(u)(\mathbf{1}\{u=v\}-p(v))$ and
$\partial D_{\mathrm{KL}}(p\|q)/\partial p(u)=\log(p(u)/q(u))+1$.
Summing over $u$ gives
\begin{align*}
\frac{\partial D_{\mathrm{KL}}(p\|q)}{\partial z_v}
&=p(v)\left(\log\frac{p(v)}{q(v)}+1\right)
-p(v)\sum_u p(u)\left(\log\frac{p(u)}{q(u)}+1\right)\\
&=p(v)\left(\log\frac{p(v)}{q(v)}-D_{\mathrm{KL}}(p\|q)\right).
\end{align*}
Multiplying by $w_{j,i}$ proves \eqref{eq:reverse-kl-logit-gradient}. Since $q_t$ is stopped, the chain rule gives $g_t=J_\ell^\top r_t$.

\partitle{Proof of Proposition~\ref{prop:absorption}}
For a differentiable function $F$ with an $M$-Lipschitz gradient in a neighborhood of the segment from $\theta$ to $\theta+\Delta$, the fundamental theorem of calculus and Cauchy--Schwarz give
\begin{align}
F(\theta+\Delta)-F(\theta)
&=\langle\nabla F(\theta),\Delta\rangle
+\int_0^1\langle\nabla F(\theta+s\Delta)-\nabla F(\theta),\Delta\rangle\,ds
\nonumber\\
&\leq\langle\nabla F(\theta),\Delta\rangle
+\frac{M}{2}\|\Delta\|_2^2.
\label{eq:descent-lemma}
\end{align}
Apply \eqref{eq:descent-lemma} at $\theta=\theta_t$ with
$F=\mathcal{L}_t$ and $\Delta=-\eta_t g_t$. Since
$\nabla\mathcal{L}_t(\theta_t)=g_t$,
\begin{align*}
\mathcal{L}_t(\theta_{t+1})
&\leq\mathcal{L}_t(\theta_t)-\eta_t\|g_t\|_2^2
+\frac{M_t\eta_t^2}{2}\|g_t\|_2^2\\
&=\mathcal{L}_t(\theta_t)
-\eta_t\left(1-\frac{M_t\eta_t}{2}\right)\|g_t\|_2^2.
\end{align*}
Since $g_t\neq0$ and $0<\eta_t<2/M_t$, $\eta_t(1-M_t\eta_t/2)\|g_t\|_2^2>0$, proving \eqref{eq:absorption-bound}.

\partitle{Proof of \eqref{eq:cross-loop-transport}}
For $d\in\{\ell,L\}$, assume $z_d$ is differentiable and its Jacobian is $C_d$-Lipschitz in operator norm in a neighborhood of the update segment. Let $J_d(\theta)=\partial z_d(\theta)/\partial\theta$, with $J_d=J_d(\theta_t)$. Integration gives
\begin{equation}
z_d(\theta_t+\Delta)-z_d(\theta_t)-J_d\Delta
=\int_0^1\bigl(J_d(\theta_t+s\Delta)-J_d\bigr)\Delta\,ds.
\label{eq:logit-remainder-integral}
\end{equation}
The norm of the right-hand side is at most $C_d\|\Delta\|_2^2/2$. With $\Delta=-\eta_tg_t$ and $g_t=J_\ell^\top r_t$, this gives
\begin{equation}
\left\|z_d(\theta_{t+1})-z_d(\theta_t)
+\eta_tJ_dJ_\ell^\top r_t\right\|_2
\leq\frac{C_d\eta_t^2}{2}\|g_t\|_2^2,
\label{eq:transport-remainder-bound}
\end{equation}
and hence \eqref{eq:cross-loop-transport}. The joint empirical tangent kernel has the block form
\begin{equation}
\begin{bmatrix}
J_\ell J_\ell^\top & J_\ell J_L^\top\\
J_LJ_\ell^\top & J_LJ_L^\top
\end{bmatrix}
=
\begin{bmatrix}J_\ell\\J_L\end{bmatrix}
\begin{bmatrix}J_\ell\\J_L\end{bmatrix}^{\!\top}.
\label{eq:joint-kernel-blocks}
\end{equation}
Thus, $K_{L\ell}=J_LJ_\ell^\top$ is the cross-loop block of the empirical tangent kernel~\citep{jacot2018neural}.

\partitle{Proof of Theorem~\ref{thm:joint-improvement}}
For each $d\in\{\ell,L\}$, apply \eqref{eq:descent-lemma} at $\theta=\theta_t$ with $F=\mathcal{R}_d$ and $\Delta=-\eta_t g_t$. This gives
\begin{equation}
\mathcal{R}_d(\theta_{t+1})-\mathcal{R}_d(\theta_t)
\leq
-\eta_t\gamma_{d,t}
+\frac{\beta_d\eta_t^2}{2}\|g_t\|_2^2.
\label{eq:joint-risk-bound}
\end{equation}
Since $\gamma_{\ell,t}>0$ and $\gamma_{L,t}>0$, the step-size condition in \eqref{eq:joint-step-size} makes the right-hand side strictly negative for both $d\in\{\ell,L\}$, proving Theorem~\ref{thm:joint-improvement}.

\partitle{Proof of \eqref{eq:evaluation-transport}}
Let $\widetilde z_d$ concatenate the logits on the fixed evaluation set $\mathcal{E}$, and let $\widetilde J_d=\partial\widetilde z_d/\partial\theta|_{\theta_t}$ and $a_{d,t}=\nabla_{\widetilde z_d}\mathcal{R}_d|_{\theta_t}$.
By the chain rule,
\begin{equation}
\gamma_{d,t}
=
\bigl(\widetilde J_d^\top a_{d,t}\bigr)^\top J_\ell^\top r_t
=
a_{d,t}^\top\widetilde J_dJ_\ell^\top r_t,
\qquad d\in\{\ell,L\},
\label{eq:evaluation-alignment-proof}
\end{equation}
which proves \eqref{eq:evaluation-transport}.

\partitle{Cross-Loop Gradient Alignment}
Consider binary next token prediction with a shared recurrent map and output head. Let token $1$ be correct, with logits $(a,0)$ and probability $\sigma(a)=(1+e^{-a})^{-1}$. In both examples, we use a single fixed prefix $c$ with unit weight, only the scalar parameter $\theta$ is trainable, and take
\[
\mathcal{R}_d(\theta)
=
-\log P_\theta^{(d)}(1\mid c).
\]
For a positively aligned case, let $h^{(0)}=0$, $h^{(k)}=h^{(k-1)}+\theta$, and use logits $(h^{(d)},0)$. Take $\ell=2$, $L=4$, and $\theta_t>0$. Then $p=\sigma(2\theta_t)$ and $q=\sigma(4\theta_t)>p$. The stopped-target reverse-KL gradient is
\begin{equation}
g_t
=
2p(1-p)(2\theta_t-4\theta_t)
<0.
\label{eq:positive-recurrent-example}
\end{equation}
Thus a gradient step increases $\theta$, increasing the correct token probabilities at both depths and decreasing both negative log-likelihoods.
For a contrasting case, consider
\begin{equation}
h^{(0)}=\begin{bmatrix}0\\0\end{bmatrix},\qquad
h^{(k)}=Ah^{(k-1)}+\theta\begin{bmatrix}1\\0\end{bmatrix},\qquad
A=\begin{bmatrix}0&-1\\1&-1\end{bmatrix},
\label{eq:counterexample-recurrence}
\end{equation}
with the shared head $g(h)=((1,-2)h,0)$. Direct substitution gives $h^{(2)}=(\theta,\theta)^\top$ and $h^{(4)}=(\theta,0)^\top$, so the correct token logits are $-\theta$ at loop $2$ and $\theta$ at loop $4$. At $\theta_t=1$, $p=\sigma(-1)<q=\sigma(1)$. The stopped-target gradient is
\begin{equation}
g_t
=
-p(1-p)(-1-1)
=
2p(1-p)>0,
\qquad
\theta_{t+1}=1-\eta_tg_t<1.
\label{eq:counterexample-update}
\end{equation}
The student probability $\sigma(-\theta)$ increases while the terminal probability $\sigma(\theta)$ decreases. Moreover, $\mathcal{R}'_\ell(1)=1-p>0$ and $\mathcal{R}'_L(1)=q-1<0$, giving $\gamma_{\ell,t}>0$ and $\gamma_{L,t}<0$. Thus, a stronger terminal teacher does not by itself imply $\gamma_{L,t}>0$.

\section{Algorithms}
\label{app:algorithms}
Algorithms~\ref{alg:loopopd} and~\ref{alg:d-loopopd} describe the training procedures in Sections~\ref{sec:loopopd} and~\ref{sec:dynamic-loopopd}. Each iteration samples $G$ responses for each of $B$ prompts, giving $n_t=BG$ responses. Teacher and student distributions are computed on the same student generated prefixes. $\operatorname{OptimizerStep}$ denotes one update with the chosen optimizer.

\clearpage

\begin{algorithm}[H]
\caption{LoopOPD: Cross-Loop On-Policy Distillation}
\label{alg:loopopd}
\small
\begin{algorithmic}[1]
\Require Training prompts $\mathcal{D}$; initial parameters $\theta_0$; student depth $\ell<L$; terminal depth $L$; prompt batch size $B$; rollouts per prompt $G$; number of updates $T$; learning rates $\{\eta_t\}_{t=0}^{T-1}$; optimizer.
\Ensure Trained parameters $\theta_T$, defining
$\pi_{\theta_T}^{(\ell)}$ and $\pi_{\theta_T}^{(L)}$.
\State Initialize the trainable LoopLM with $\theta_0$.
\State Create a frozen teacher copy with $\bar\theta_0\gets\theta_0$.
\For{$t=0,\ldots,T-1$}
    \State \textbf{Student rollouts}
    \State Draw $B$ prompts $x_1,\ldots,x_B$ from $\mathcal{D}$.
    \For{$b=1,\ldots,B$}
        \State Sample $y^{(b,s)}\sim\pi_{\theta_t}^{(\ell)}(\cdot\mid x_b)$,
        $s=1,\ldots,G$.
    \EndFor
    \State Flatten the responses into
    $\mathcal{B}_t=\{(x^{(j)},y^{(j)})\}_{j=1}^{n_t}$, where $n_t=BG$.
    \State Set $m_j\gets |y^{(j)}|$ and
    $c_{j,i}\gets(x^{(j)},y_{<i}^{(j)})$ for $j=1,\ldots,n_t$,
    $i=1,\ldots,m_j$.
    \State \textbf{Frozen teacher targets}
    \State Evaluate the frozen teacher at depth $L$ on the batch prefixes.
    \State $q_{j,i}(v)\gets
    P_{\bar\theta_0}^{(L)}(v\mid c_{j,i})$ for all $(j,i)$ and
    $v\in\mathcal{V}$.
    \State Evaluate the trainable LoopLM at depth $\ell$ on the same prefixes.
    \State $p_{j,i}(v)\gets P_{\theta_t}^{(\ell)}(v\mid c_{j,i})$
    for all $(j,i)$ and $v\in\mathcal{V}$.
    \State $\displaystyle
    \kappa_{j,i}\gets\sum_{v\in\mathcal{V}}p_{j,i}(v)
    \log\frac{p_{j,i}(v)}{q_{j,i}(v)}$ for all $(j,i)$.
    \State $\displaystyle
    \mathcal{L}_t\gets\frac{1}{n_t}\sum_{j=1}^{n_t}
    \frac{1}{m_j}\sum_{i=1}^{m_j}\kappa_{j,i}$.
    \State $g_t\gets\left.\nabla_\theta\mathcal{L}_t\right|_{\theta=\theta_t}$.
    \State $\theta_{t+1}\gets
    \operatorname{OptimizerStep}(\theta_t,g_t,\eta_t)$.
\EndFor
\State \Return $\theta_T$.
\end{algorithmic}
\end{algorithm}

\begin{algorithm}[H]
\caption{D-LoopOPD: Dynamic Cross-Loop On-Policy Distillation}
\label{alg:d-loopopd}
\small
\begin{algorithmic}[1]
\Require Training prompts $\mathcal{D}$; initial parameters $\theta_0$; student depth $\ell<L$; terminal depth $L$; prompt batch size $B$; rollouts per prompt $G$; number of updates $T$; learning rates $\{\eta_t\}_{t=0}^{T-1}$; optimizer.
\Ensure Trained parameters $\theta_T$, defining
$\pi_{\theta_T}^{(\ell)}$ and $\pi_{\theta_T}^{(L)}$.
\State Initialize the trainable LoopLM with $\theta_0$.
\For{$t=0,\ldots,T-1$}
    \State \textbf{Student rollouts}
    \State Draw $B$ prompts $x_1,\ldots,x_B$ from $\mathcal{D}$.
    \For{$b=1,\ldots,B$}
        \State Sample $y^{(b,s)}\sim\pi_{\theta_t}^{(\ell)}(\cdot\mid x_b)$,
        $s=1,\ldots,G$.
    \EndFor
    \State Flatten the responses into
    $\mathcal{B}_t=\{(x^{(j)},y^{(j)})\}_{j=1}^{n_t}$, where $n_t=BG$.
    \State Set $m_j\gets |y^{(j)}|$ and
    $c_{j,i}\gets(x^{(j)},y_{<i}^{(j)})$ for $j=1,\ldots,n_t$,
    $i=1,\ldots,m_j$.
    \State \textbf{Current teacher targets}
    \State Evaluate the trainable LoopLM at depth $L$ on the batch prefixes.
    \State $q_{j,i}(v)\gets
    \operatorname{sg}\!\left[P_{\theta_t}^{(L)}(v\mid c_{j,i})\right]$
    for all $(j,i)$ and $v\in\mathcal{V}$.
    \State Evaluate the trainable LoopLM at depth $\ell$ on the same prefixes.
    \State $p_{j,i}(v)\gets P_{\theta_t}^{(\ell)}(v\mid c_{j,i})$
    for all $(j,i)$ and $v\in\mathcal{V}$.
    \State $\displaystyle
    \kappa_{j,i}\gets\sum_{v\in\mathcal{V}}p_{j,i}(v)
    \log\frac{p_{j,i}(v)}{q_{j,i}(v)}$ for all $(j,i)$.
    \State $\displaystyle
    \mathcal{L}_t\gets\frac{1}{n_t}\sum_{j=1}^{n_t}
    \frac{1}{m_j}\sum_{i=1}^{m_j}\kappa_{j,i}$.
    \State $g_t\gets\left.\nabla_\theta\mathcal{L}_t\right|_{\theta=\theta_t}$.
    \State $\theta_{t+1}\gets
    \operatorname{OptimizerStep}(\theta_t,g_t,\eta_t)$.
    \State Use $\pi_{\theta_{t+1}}^{(L)}$ as the teacher for the next iteration.
\EndFor
\State \Return $\theta_T$.
\end{algorithmic}
\end{algorithm}

\clearpage

\section{Supplementary Experiments}
\label{app:supp-experiments}
\subsection{Additional Training Dynamics}
\label{app:additional-training-dynamics}

\partitle{Validation Performance}
Table~\ref{tab:training-validation} reports the exact AIME 2024 validation Avg@16 values corresponding to Figure~\ref{fig:training_dynamics} (a) for Ouro-2.6B-Thinking at the student loop $\ell=2$.
\begin{table}[h]
\caption{AIME 2024 validation Avg@16 during training for Ouro-2.6B-Thinking at student loop $\ell=2$. Best results at each training step are shown in \textbf{bold}.}
\label{tab:training-validation}
\begin{center}
\begin{tabular}{c|ccccc}
\toprule
\textbf{Method} & \textbf{Step 20} & \textbf{Step 40} & \textbf{Step 60} & \textbf{Step 80} & \textbf{Step 100} \\
\midrule
LoopOPD
& 39.37 & 40.00 & 42.08 & 40.62 & 45.62 \\
D-LoopOPD
& \textbf{41.46} & \textbf{40.20} & \textbf{46.25} & \textbf{44.79} & \textbf{47.50} \\
\bottomrule
\end{tabular}
\end{center}
\end{table}

\begin{figure}[h]
    \centering
    \includegraphics[width=0.49\linewidth]{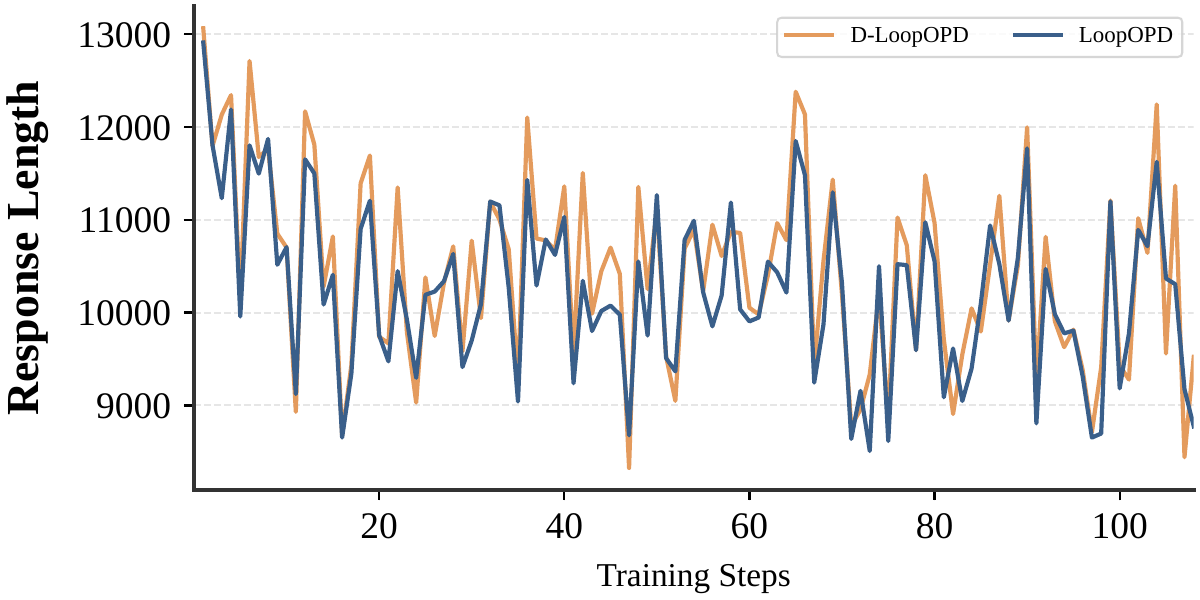}
    \hfill
    \includegraphics[width=0.49\linewidth]{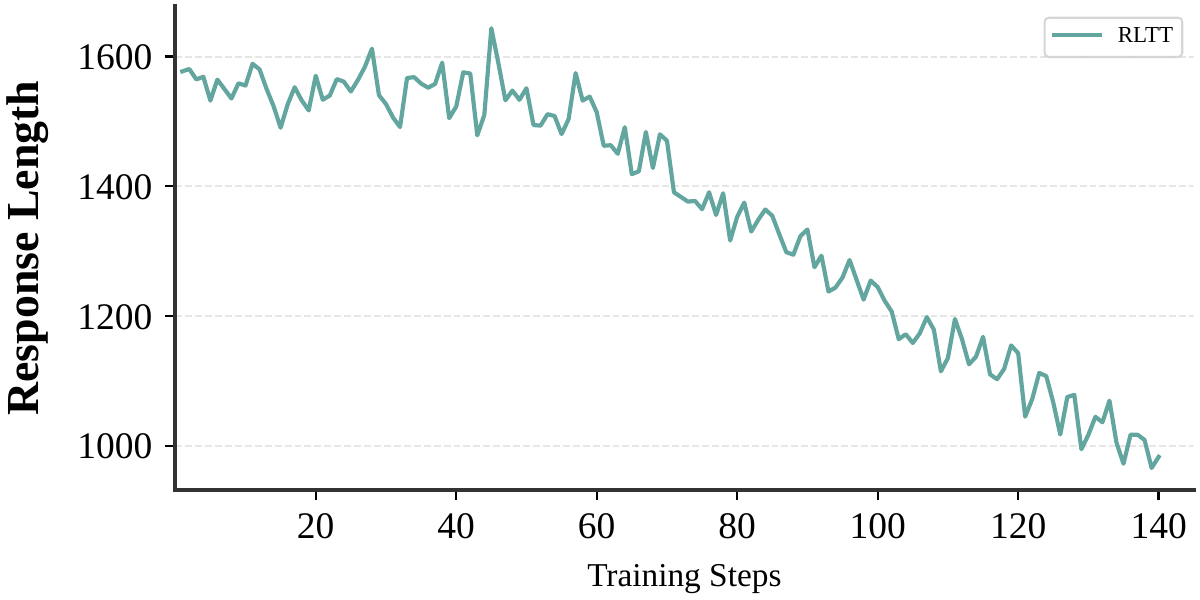}
    \caption{Response length during training for Ouro-2.6B-Thinking. \textbf{Left:} LoopOPD and D-LoopOPD. \textbf{Right:} RLTT. LoopOPD and D-LoopOPD use a maximum rollout length of 31,744 tokens, while RLTT uses 2,048 tokens.}
    \label{fig:response-length}
\end{figure}

\begin{figure}[h]
    \centering
    \includegraphics[width=0.68\linewidth]{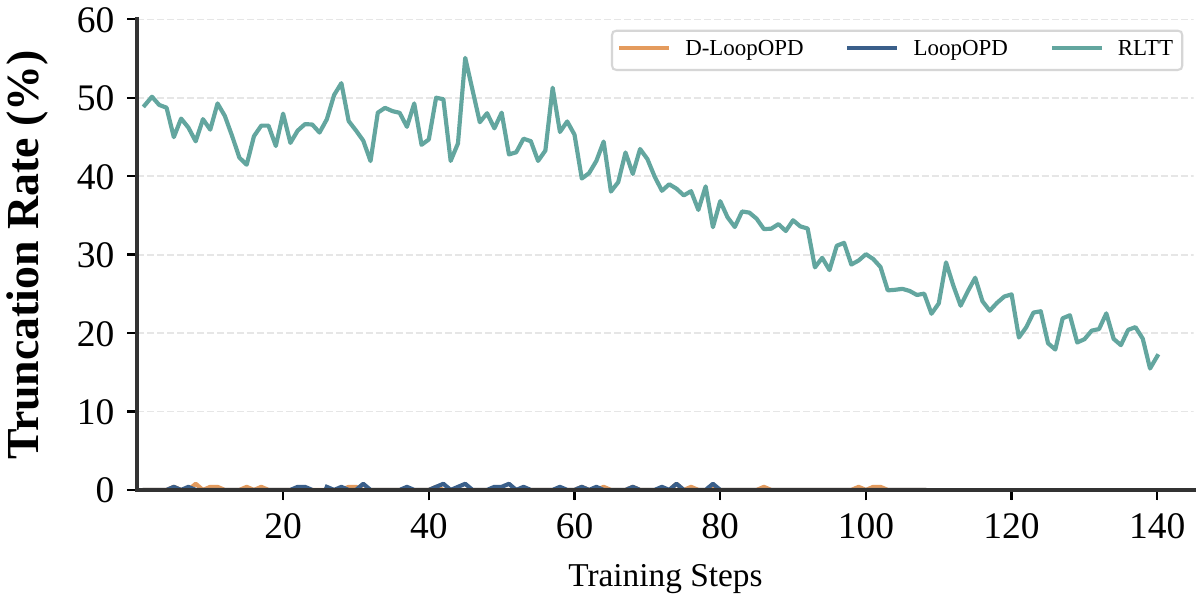}
    \caption{Rollout truncation rate during training for Ouro-2.6B-Thinking.
    RLTT exhibits substantially higher truncation than LoopOPD and D-LoopOPD throughout training.}
    \label{fig:truncation-rate}
\end{figure}

\partitle{Response Length and Truncation Rate}
Figure~\ref{fig:response-length} compares response length during training, while Figure~\ref{fig:truncation-rate} reports the rollout truncation rate.
Due to the GPU memory overhead of retaining learning signals across recurrent loops, we follow the original RLTT training configuration and use a maximum rollout length of 2,048 tokens for RLTT, while LoopOPD and D-LoopOPD use 31,744 tokens.
RLTT produces substantially shorter responses and exhibits a substantially higher truncation rate throughout training. In contrast, LoopOPD and D-LoopOPD generate much longer responses while maintaining near zero truncation. This behavior is consistent with the decode budget results in Appendix~\ref{app:decode-budget}, where RLTT performs strongly under short decode budgets but saturates as the budget increases, while D-LoopOPD continues to benefit from longer generation. Training RLTT with 31,744 token rollouts exceeded our available memory budget, highlighting the practical limitation of extending per loop reward supervision to long rollouts.

\subsection{Cross-Loop Performance}
\label{app:cross-loop-performance}
Table~\ref{tab:cross-loop-results} reports the exact values corresponding to the cross-loop results shown in Figure~\ref{fig:training_dynamics} (b). We evaluate each loop depth of Ouro-2.6B-Thinking on AIME 2024 using a maximum response length of 31,744 tokens. Each entry reports Avg@16 / Pass@16.
\begin{table*}[h]
\caption{Cross-loop performance on AIME 2024 for Ouro-2.6B-Thinking. Best results at each loop depth are shown in \textbf{bold}.}
\label{tab:cross-loop-results}
\begin{center}
\resizebox{0.8\textwidth}{!}{
\begin{tabular}{lcccc}
\toprule
\textbf{Method}
& \textbf{Loop 1}
& \textbf{Loop 2}
& \textbf{Loop 3}
& \textbf{Loop 4} \\
\midrule

Base
& 0.63 / 3.33
& 39.16 / 73.33
& 61.46 / 83.33
& 62.08 / 83.33 \\

RLTT
& 0.00 / 0.00
& 26.87 / 63.33
& 49.58 / 80.00
& 56.67 / 83.33 \\

GRPO
& 0.42 / 3.33
& 41.88 / 73.33
& 62.92 / 83.33
& 62.92 / 86.67 \\

\cellcolor{oursbg}LoopOPD
& \cellcolor{oursbg}1.25 / \textbf{10.00}
& \cellcolor{oursbg}45.62 / 76.67
& \cellcolor{oursbg}63.33 / 83.33
& \cellcolor{oursbg}65.21 / \textbf{90.00} \\

\cellcolor{oursbg}D-LoopOPD
& \cellcolor{oursbg}\textbf{1.88} / \textbf{10.00}
& \cellcolor{oursbg}\textbf{47.50} / \textbf{80.00}
& \cellcolor{oursbg}\textbf{66.87} / \textbf{86.67}
& \cellcolor{oursbg}\textbf{71.67} / \textbf{90.00} \\

\bottomrule
\end{tabular}
}
\end{center}
\end{table*}

\subsection{Effect of Decode Budget}
\label{app:decode-budget}
\begin{figure}[h]
    \centering
    \includegraphics[width=0.8\linewidth]{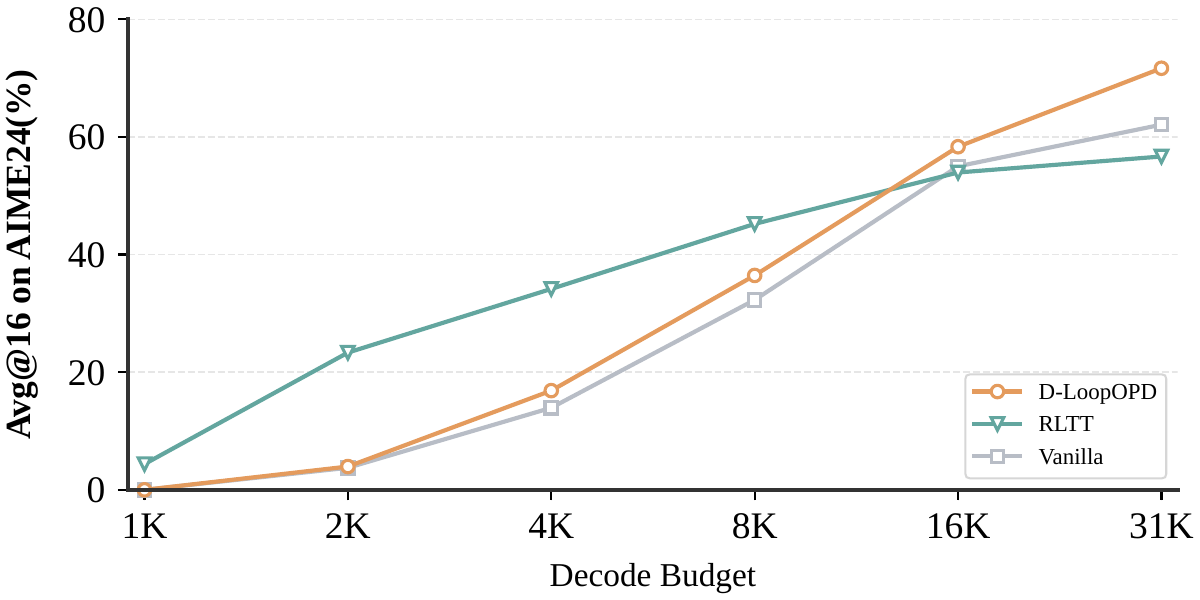}
    \caption{Effect of decode budget on AIME 2024 performance for Ouro-2.6B-Thinking.
    Avg@16 is evaluated under maximum decode budgets ranging from 1K to 31K tokens.}
    \label{fig:decode-budget}
\end{figure}

Figure~\ref{fig:decode-budget} shows AIME 2024 Avg@16 of Ouro-2.6B-Thinking under different maximum decode budgets, with exact values reported in Table~\ref{tab:decode-budget}. RLTT performs better under short decode budgets, but its gains diminish as the budget increases. D-LoopOPD overtakes RLTT at 16K tokens and continues to improve at 31K, reaching 71.67 Avg@16 compared with 56.67 for RLTT. Notably, RLTT also falls below the vanilla model at both 16K and 31K, indicating a degradation in its ability to benefit from longer reasoning trajectories. In contrast, D-LoopOPD exceeds the vanilla model across all decode budgets and outperforms it from 2K onward, suggesting that its gains do not come at the cost of long budget reasoning capability. These results are consistent with the response length and truncation behavior in Figures~\ref{fig:response-length} and~\ref{fig:truncation-rate}: RLTT favors short generation under its constrained rollout budget, whereas D-LoopOPD continues to benefit from longer generation.

\begin{table}[h]
\caption{AIME 2024 Avg@16 under different maximum decode budgets for Ouro-2.6B-Thinking. Best results at each decode budget are shown in \textbf{bold}.}
\label{tab:decode-budget}
\begin{center}
\begin{tabular}{c|cccccc}
\toprule
\textbf{Method} & \textbf{1K} & \textbf{2K} & \textbf{4K} & \textbf{8K} & \textbf{16K} & \textbf{31K} \\
\midrule
Base
& 0.00 & 3.75 & 13.95 & 32.29 & 55.00 & 62.08 \\
RLTT
& \textbf{4.38} & \textbf{23.33} & \textbf{34.16} & \textbf{45.21} & 53.95 & 56.67 \\
D-LoopOPD
& 0.00 & 3.96 & 16.87 & 36.45 & \textbf{58.33} & \textbf{71.67} \\
\bottomrule
\end{tabular}
\end{center}
\end{table}

\section{Implementation Details}
\label{app:details}

\subsection{Training Configuration}
We detail the training configurations used throughout our experiments in Table~\ref{tab:training-hyperparameters}. All post-training methods use DAPO-Math-17k as the training dataset, with its basic statistics summarized in Table~\ref{tab:training-data}. Each training run uses 4 NVIDIA H200 GPUs. For OPSD, we follow the original setup, using a maximum sequence length of 20,000 and a maximum rollout length of 1,024 tokens.
RLTT uses a maximum rollout length of 2,048 tokens and AdamW8bit, following its original training configuration, consistent with the higher GPU memory usage associated with retaining learning signals across recurrent loops. We use uniform weights across recurrent loops.
LoopOPD and D-LoopOPD use student depth $\ell=2$ and terminal depth $L=4$.
\begin{table*}[h]
\caption{Training and validation hyperparameters for different methods.}
\label{tab:training-hyperparameters}
\begin{center}
\resizebox{\textwidth}{!}{
\begin{tabular}{lcccccc}
\toprule
\textbf{Hyperparameter}
& \textbf{GRPO}
& \textbf{RLTT}
& \textbf{OPD}
& \textbf{OPSD}
& \textbf{LoopOPD}
& \textbf{D-LoopOPD} \\
\midrule

\rowcolor{gray!10}
\multicolumn{7}{l}{\textbf{\textit{General}}} \\
Max Prompt Length
& 1024 & 1024 & 1024 & -- & 1024 & 1024 \\

\midrule
\rowcolor{gray!10}
\multicolumn{7}{l}{\textbf{\textit{Optimization}}} \\
Global Batch Size
& 256 & 512 & 256 & 32 & 256 & 256 \\
Prompts per Rollout Batch
& 32 & 32 & 32 & 32 & 32 & 32 \\
Group Rollout Size
& 8 & 8 & 8 & 1 & 8 & 8 \\
Gradient Accumulation
& 1 & 2 & 1 & 1 & 1 & 1 \\
Optimizer
& AdamW & AdamW8bit & AdamW & AdamW & AdamW & AdamW \\
Learning Rate
& $1\times10^{-6}$
& $1\times10^{-6}$
& $1\times10^{-6}$
& $5\times10^{-6}$
& $1\times10^{-6}$
& $1\times10^{-6}$ \\
Warmup
& 10 steps & 10\% & -- & -- & -- & 10 steps \\
Weight Decay
& 0.1 & 0.1 & 0.01 & 0 & 0.01 & 0.01 \\

\midrule
\rowcolor{gray!10}
\multicolumn{7}{l}{\textbf{\textit{Sampling}}} \\
Rollout Max Response Length
& 31744 & 2048 & 31744 & 1024 & 31744 & 31744 \\
Rollout Temperature
& 1.0 & 1.0 & 1.0 & 1.1 & 1.0 & 1.0 \\
Rollout Top-$p$
& 1.0 & 1.0 & 1.0 & 0.95 & 1.0 & 1.0 \\
Validation Max Response Length
& 31744 & 31744 & 31744 & 31744 & 31744 & 31744 \\
Validation Temperature
& 1.0 & 1.0 & 1.0 & 1.0 & 1.0 & 1.0 \\
Validation Top-$p$
& 0.7 & 0.7 & 0.7 & 0.7 & 0.7 & 0.7 \\
Validation Samples per Problem
& 16 & 16 & 16 & 16 & 16 & 16 \\

\midrule
\rowcolor{gray!10}
\multicolumn{7}{l}{\textbf{\textit{Method Specific}}} \\
Student Depth $\ell$
& -- & -- & 4 & 4 & 2 & 2 \\
Teacher Depth $L$
& -- & -- & 4 & 4 & 4 & 4 \\
Frozen Teacher
& -- & -- & Yes & Yes & Yes & No \\
External Teacher
& -- & -- & Yes & No & No & No \\
Privileged Information
& -- & -- & No & Yes & No & No \\

\bottomrule
\end{tabular}
}
\end{center}
\end{table*}

\begin{table}[t]
\caption{Training dataset used for post-training.}
\label{tab:training-data}
\begin{center}

\resizebox{0.85\linewidth}{!}{
\begin{tabular}{lcccc}
\toprule
\textbf{Dataset}
& \textbf{Split}
& \textbf{\# Examples}
& \textbf{Domain}
& \textbf{Answer Type} \\
\midrule
DAPO-Math-17k
& Train
& 17,398
& Mathematical Reasoning
& Integer Ground Truth \\
\bottomrule
\end{tabular}
}
\end{center}
\end{table}

\subsection{Evaluation Protocol}
Table~\ref{tab:evaluation-details} summarizes the evaluation protocols used throughout our experiments. For mathematical reasoning benchmarks, we use zero-shot generation with thinking enabled. For MMLU, we use the standard 5-shot multiple-choice log-likelihood evaluation, while code generation benchmarks are evaluated using EvalPlus.

\partitle{Mathematical Reasoning}
For all mathematical reasoning benchmarks, each example is provided as a single user message containing the original problem followed by:
\begin{tcolorbox}[
    enhanced,
    colback=white,
    colframe=gray!65,
    colbacktitle=gray!65,
    coltitle=white,
    fonttitle=\bfseries,
    title={Mathematical Reasoning Prompt},
    boxrule=0.8pt,
    arc=2mm,
    outer arc=2mm,
    titlerule=0pt,
    left=8pt,
    right=8pt,
    top=8pt,
    bottom=8pt
]
\ttfamily
\{problem\}

Please reason step by step, and put your final answer within
\textbackslash boxed\{\}.
\end{tcolorbox}
For AIME 2024, AIME 2025, and AIME 2026, we evaluate all 30 problems from each benchmark under the same protocol. Each problem is evaluated with 16 sampled completions using temperature $1.0$, top-$p$ $0.7$, and a maximum of 31,744 generated tokens. We extract the final boxed answer and determine correctness using numerical and symbolic mathematical equivalence. We report two metrics. Avg@16 is the fraction of correct completions among the $30\times16=480$ generated responses for each benchmark. Pass@16 is the fraction of the 30 problems for which at least one of the 16 completions is correct.

For OlympiadBench, we evaluate 675 text-based mathematical problems. For BeyondAIME, we use the official test split containing 100 problems. Both benchmarks use temperature $1.0$, top-$p$ $0.7$, and a maximum of 31,744 generated tokens.

\partitle{MMLU}
We evaluate MMLU using \texttt{lm-eval-harness} on the full test split of 14,042 questions across 57 subjects. We use five demonstrations from the development split of each subject together with the subject description. The prompt is formatted as:
\begin{tcolorbox}[
    enhanced,
    colback=white,
    colframe=gray!65,
    colbacktitle=gray!65,
    coltitle=white,
    fonttitle=\bfseries,
    title={MMLU Prompt},
    boxrule=0.8pt,
    arc=2mm,
    outer arc=2mm,
    titlerule=0pt,
    left=8pt,
    right=8pt,
    top=8pt,
    bottom=8pt
]
\ttfamily
\{subject description\}

\{five demonstrations\}

\{question\}

A. \{choice\_0\}

B. \{choice\_1\}

C. \{choice\_2\}

D. \{choice\_3\}

Answer:
\end{tcolorbox}
We select the candidate with the highest log-likelihood among A, B, C, and D.

\partitle{Code Generation}
We evaluate MBPP, MBPP+, HumanEval, and HumanEval+ using EvalPlus. MBPP and MBPP+ contain 378 problems, while HumanEval and HumanEval+ contain 164 problems. For each problem, we generate one program and use it to compute both the original and ``+'' benchmark scores. The original benchmark uses the base test suite, while the ``+'' variant additionally includes the extended tests. We use greedy decoding with one generated solution per problem and a maximum of 768 new tokens.
The prompt is formatted as:
\begin{tcolorbox}[
    enhanced,
    colback=white,
    colframe=gray!65,
    colbacktitle=gray!65,
    coltitle=white,
    fonttitle=\bfseries,
    title={Code Generation Prompt},
    boxrule=0.8pt,
    arc=2mm,
    outer arc=2mm,
    titlerule=0pt,
    left=8pt,
    right=8pt,
    top=8pt,
    bottom=8pt
]
\ttfamily
User:

Please provide a self-contained Python script that solves the following problem in a markdown code block:

\texttt{```}

\{task prompt\}

\texttt{```}

Assistant:

Below is a Python script with a self-contained function that solves the problem and passes corresponding tests:

\texttt{```python}
\end{tcolorbox}

\begin{table*}[h]
\caption{Evaluation settings for different benchmarks.}
\label{tab:evaluation-details}
\begin{center}
\small
\begin{tabular}{lcccc}
\toprule
\textbf{Benchmark} & \textbf{\# Examples} & \textbf{Decoding / Scoring} & \textbf{Max New Tokens} & \textbf{Metric} \\
\midrule
AIME 2024 & 30 & $T=1.0$, $p=0.7$, $n=16$ & 31,744 & Avg@16 / Pass@16 \\
AIME 2025 & 30 & $T=1.0$, $p=0.7$, $n=16$ & 31,744 & Avg@16 / Pass@16 \\
AIME 2026 & 30 & $T=1.0$, $p=0.7$, $n=16$ & 31,744 & Avg@16 / Pass@16 \\
OlympiadBench & 675 & $T=1.0$, $p=0.7$, $n=1$ & 31,744 & Accuracy \\
BeyondAIME & 100 & $T=1.0$, $p=0.7$, $n=1$ & 31,744 & Accuracy \\
\midrule
MMLU & 14,042 & Candidate log-likelihood & -- & Accuracy \\
MBPP & 378 & Greedy, $n=1$ & 768 & Pass@1 \\
MBPP+ & 378 & Greedy, $n=1$ & 768 & Pass@1 \\
HumanEval & 164 & Greedy, $n=1$ & 768 & Pass@1 \\
HumanEval+ & 164 & Greedy, $n=1$ & 768 & Pass@1 \\
\bottomrule
\end{tabular}
\end{center}
\end{table*}

\section{Case Study}
\label{app:case-study}
\newcommand{\BaseError}[1]{\textcolor{red!75!black}{#1}}
\newcommand{\BaseNote}[1]{%
\par\smallskip
{\small\color{red!75!black}\textit{Annotation: #1}}
\par\smallskip
}
We present a qualitative case study to compare the reasoning behaviors of the vanilla model and the D-LoopOPD trained model.
\begin{tcolorbox}[
    enhanced,
    breakable,
    colback=white,
    colframe=gray!65,
    colbacktitle=gray!65,
    coltitle=white,
    fonttitle=\bfseries,
    title={Question and Ground Truth},
    boxrule=0.8pt,
    arc=2mm,
    outer arc=2mm,
    titlerule=0pt,
    left=8pt,
    right=8pt,
    top=8pt,
    bottom=8pt
]
\textbf{Question:} 
Let \(b\ge 2\) be an integer. Call a positive integer \(n\) \(b\text{-}\mathit{eautiful}\) if it has exactly two digits when expressed in base \(b\) and these two digits sum to \(\sqrt n\). For example, \(81\) is \(13\text{-}\mathit{eautiful}\) because \(81=\underline{6}\;\underline{3}_{13}\) and \(6+3=\sqrt{81}\). Find the least integer \(b\ge 2\) for which there are more than ten \(b\text{-}\mathit{eautiful}\) integers.

\medskip
\textbf{Ground Truth:} \(211\)
\end{tcolorbox}

\begin{tcolorbox}[
    enhanced,
    breakable,
    colback=blue!3,
    colframe=blue!30,
    colbacktitle=blue!18,
    coltitle=black,
    fonttitle=\bfseries,
    title={Response from Vanilla Model},
    boxrule=0.8pt,
    arc=2mm,
    outer arc=2mm,
    titlerule=0pt,
    left=8pt,
    right=8pt,
    top=8pt,
    bottom=8pt
]

Okay, so I need to find the smallest integer \(b\) (where \(b\) is at least 2) such that there are more than ten \(b\)-eautiful numbers. A \(b\)-eautiful number is a two-digit number in base \(b\) where the sum of its digits equals the square root of the number itself. The example given is 81 in base 13, which is written as 63 in base 13, and \(6 + 3 = 9\), which is \(\sqrt{81}\).
\par\medskip
Let me formalize this. For each base \(b\), we need to find all pairs of digits \((a,c)\) where \(a\) is between 1 and \(b-1\) (since it's the leading digit in a two-digit number), and \(c\) is between 0 and \(b-1\). Then, the number \(n\) is \(a\cdot b + c\), and we require that \(a + c = \sqrt{n}\).
\par\medskip
So, substituting \(n\) into the equation: \(a + c = \sqrt{a\cdot b + c}\). Let's square both sides to eliminate the square root: \((a + c)^2 = a\cdot b + c\).
\par\medskip
Expanding the left side: \(a^2 + 2ac + c^2 = a\cdot b + c\).
\par\medskip
Rearranging terms: \(a^2 + 2ac + c^2 - a\cdot b - c = 0\).
\par\medskip
Hmm, this seems a bit complicated. Maybe there's a better way to approach this. Let's see.
\par\medskip
Alternatively, since \(a\) and \(c\) are digits in base \(b\), their possible values are limited. For a given base \(b\), \(a\) can be from 1 to \(b-1\), and \(c\) can be from 0 to \(b-1\). So, for each base \(b\), we can iterate over all possible \(a\) and \(c\), compute \(n = a\cdot b + c\), check if \(a + c\) equals \(\sqrt{n}\), and count how many such numbers exist. Then, we need to find the smallest \(b\) where this count exceeds 10.
\par\medskip
But since we need to find the minimal \(b\) where the count is more than ten, maybe there's a pattern or formula we can derive instead of brute-forcing for each \(b\). Let's see.
\par\medskip
Starting with the equation: \((a + c)^2 = a\cdot b + c\). Let's denote \(s = a + c\). Then, \(s^2 = a\cdot b + c\). But since \(s = a + c\), we can express \(c\) as \(s - a\). Substituting into the equation: \(s^2 = a\cdot b + (s - a)\). Simplify this: \(s^2 = a\cdot b + s - a\). Rearranging terms: \(s^2 - s = a\cdot (b - 1)\). Therefore, \(a = (s^2 - s)/(b - 1)\).
\par\medskip
Since \(a\) must be a positive integer between 1 and \(b-1\), the right-hand side must also be an integer in that range. So, \((s^2 - s)\) must be divisible by \((b - 1)\), and the result must be between 1 and \(b-1\).
\par\medskip
So, for each possible \(s\), we can compute \(a\) and check if it's valid. Then, \(c = s - a\), which must be between 0 and \(b-1\).
\par\medskip
Wait, but in the example given, 81 is a \(b\)-eautiful number where \(b=13\). \(\sqrt{81}=9\), which is equal to the sum of digits \(6 + 3\). So, in that case, \(s=9\), which is equal to \(\sqrt{n}\). So, \(s\) must be an integer such that \(s^2\) is a two-digit number in base \(b\). So, \(s^2\) must be between \(b\) and \(b^2 - 1\). Therefore, \(s\) must satisfy \(\sqrt{b} \leq s \leq b - 1\). Because \(s^2 \leq b^2 - 1\) implies \(s \leq b - 1\) (since \(s\) is integer). Therefore, \(s\) ranges from \(\lceil\sqrt{b}\rceil\) to \(b - 1\).
\par\medskip
Therefore, for each base \(b\), \(s\) can be from \(\lceil\sqrt{b}\rceil\) to \(b - 1\). For each such \(s\), we can compute \(a = (s^2 - s)/(b - 1)\). Then, check if \(a\) is an integer between 1 and \(b - 1\), and \(c = s - a\) is between 0 and \(b - 1\).
\par\medskip
So, the number of \(b\)-eautiful numbers is the number of \(s\) in \([\lceil\sqrt{b}\rceil, b - 1]\) such that \((s^2 - s)\) is divisible by \((b - 1)\), and the resulting \(a\) and \(c\) are valid digits.
\par\medskip
...
\par\medskip
\(b=7\): \(s_{\min}=\lceil\sqrt{7}\rceil=3\); \(s_{\max}=6\); \(s=3, 4, 5, 6\).
\par\medskip
\(s=3\): \(\mathrm{numerator}=9-3=6\); \(b-1=6\); \(6/6=1\); \(a=1\); \(c=3 -1=2\); Check \(a=1\) valid, \(c=2\) valid. \(n=1\cdot 7 +2=9\). Check \(1+2=3=\sqrt{9}\). Yes. Count \(=1\).
\par\medskip
\(s=4\): \(\mathrm{numerator}=16-4=12\); \(12/6=2\); \(a=2\); \(c=4 -2=2\); Check \(a=2\) valid, \(c=2\) valid. \(n=2\cdot 7 +2=16\). Check \(2+2=4=\sqrt{16}\). Yes. Count \(=2\).
\par\medskip
\(s=5\): \(\mathrm{numerator}=25-5=20\); \(20/6\approx 3.333\), invalid.
\par\medskip
\(s=6\): \(\mathrm{numerator}=36-6=30\); \(30/6=5\); \(a=5\); \(c=6 -5=1\); Check \(a=5\) valid, \(c=1\) valid. \(n=5\cdot 7 +1=36\). Check \(5+1=6=\sqrt{36}\). Yes. Count \(=3\). So, for \(b=7\), count \(=3\).
\par\medskip
...
\par\medskip
Let's denote \(k = b -1\). Then, we need \((s^2 - s)\) divisible by \(k\). So, \(k\) divides \(s(s -1)\). Therefore, \(k\) must be a divisor of \(s(s -1)\). Since \(s\) and \(s -1\) are consecutive integers, they are coprime. Therefore, \(k\) must be a product of divisors of \(s\) and \(s -1\). But since \(s\) and \(s -1\) are coprime, the divisors of \(k\) must be products of divisors of \(s\) and \(s -1\).
\par\medskip
Alternatively, for each \(k\), the number of \(s\) in \([\lceil\sqrt{k+1}\rceil,k]\) such that \(k\) divides \(s(s -1)\). Let's compute this for \(k=24\) (\(b=25\)): \(s\) ranges from 5 to 24. We need 24 divides \(s(s -1)\). Since \(24=8\cdot 3\), and \(s\) and \(s -1\) are coprime, \BaseError{one of them must be divisible by 8 and the other by 3.}
\BaseNote{Both 8 and 3 can divide the same number. This misses \(s=24\).}

So, either \(s \equiv 0 \pmod{8}\) and \(s \equiv 1 \pmod{3}\), or \(s \equiv 1 \pmod{8}\) and \(s \equiv 0 \pmod{3}\). Let's solve these congruences.
\par\medskip
Case 1: \(s \equiv 0 \pmod{8}\) and \(s \equiv 1 \pmod{3}\).\par\nopagebreak[4]\smallskip
Let \(s = 8m\). Then, \(8m \equiv 1 \pmod{3} \Rightarrow 2m \equiv 1 \pmod{3} \Rightarrow m \equiv 2 \pmod{3}\). So, \(m=3n +2\). Therefore, \(s=8\cdot (3n +2)=24n +16\). Within \(s \leq 24, n=0\): \(s=16\); \(n=1\): \(s=40\) (too big). So, \(s=16\).
\par\medskip
Case 2: \(s \equiv 1 \pmod{8}\) and \(s \equiv 0 \pmod{3}\).\par\nopagebreak[4]\smallskip
Let \(s=8m +1\). Then, \(8m +1 \equiv 0 \pmod{3} \Rightarrow 2m +1 \equiv 0 \pmod{3} \Rightarrow 2m \equiv -1 \equiv 2 \pmod{3} \Rightarrow m \equiv 1 \pmod{3}\). So, \(m=3n +1\). Therefore, \(s=8\cdot (3n +1) +1=24n +9\). Within \(s \leq 24, n=0\): \(s=9\); \(n=1\): \(s=33\) (too big). So, \(s=9\).
\par\medskip
Therefore, for \(k=24\), the possible \(s\) are \BaseError{9 and 16}. Then, check if these \(s\)'s give valid \(a\) and \(c\).
\par\medskip
For \(s=9\): \(a=(81 -9)/24=72/24=3\); \(c=9 -3=6\); Check \(a=3\) is between 1 and 24 (valid), \(c=6\) is between 0 and 24 (valid). So, \(n=3\cdot 25 +6=81\). Valid.
\par\medskip
For \(s=16\): \(a=(256 -16)/24=240/24=10\); \(c=16 -10=6\); Check \(a=10\) valid, \(c=6\) valid. \(n=10\cdot 25 +6=256\). Valid.
\par\medskip
...
\par\medskip
\(s=24\): \(\mathrm{numerator}=576-24=552, 552/24=23, a=23, c=1\). Valid.
\par\medskip
...
\par\medskip
Therefore, our earlier analysis missed \(s=24\) because we considered only \(s=9\) and 16. But \(s=24\) also works. Therefore, the number of solutions is more than 2.
\par\medskip
Therefore, the earlier approach of solving congruences missed some cases. So, perhaps the number of solutions is equal to the number of divisors of \(k= b -1\) that divide \(s\) or \(s -1\).
\par\medskip
Alternatively, since \(k\) divides \(s(s -1)\), and \(s\) and \(s -1\) are coprime, \BaseError{then \(k\) must divide \(s\) or \(k\) divides \(s -1\).}

Wait, no. Since \(s\) and \(s -1\) are coprime, \BaseError{\(k\) must divide \(s\) or \(k\) divides \(s -1\).} Because if a prime divides \(k\), it can divide either \(s\) or \(s -1\), but not both. Therefore, \(k\) must be a product of divisors of \(s\) and \(s -1\). But since \(s\) and \(s -1\) are coprime, \(k\) must be a product of divisors of \(s\) and divisors of \(s -1\). Therefore, the number of solutions \(s\) is equal to the number of ways to write \(k\) as a product of two coprime factors, one dividing \(s\) and the other dividing \(s -1\).
\BaseNote{For \(k=24\) and \(s=9\), \(24\mid9\cdot8\), but 24 divides neither 9 nor 8.}

Therefore, the number of solutions \(s\) is equal to \BaseError{\(2^{m-1}\)}, where \(m\) is the number of distinct prime factors of \(k\).

Because each prime factor can go to either \(s\) or \(s -1\), but since they are coprime, each prime must go entirely to one or the other. Therefore, the number of solutions is \BaseError{\(2^{m-1}\)}, but this might not be exact.
\BaseNote{There are \(2^m\) solutions modulo \(k\), and \(2^m-1\) valid values after excluding \(s=1\).}

Wait, for example, if \(k=24=2^3\cdot 3\), then the number of ways to split the prime factors into \(s\) and \(s -1\). Since \(s\) and \(s -1\) are coprime, all factors of 2 must go to one of them, \BaseError{and all factors of 3 must go to the other.} So, either \(s\) is divisible by \(2^3\) and 3, and \BaseError{\(s -1\) is 1, which is impossible since \(s \geq 2\).} Or, \(s\) is divisible by \(2^3\) and \(s -1\) is divisible by 3. Or, \(s\) is divisible by 3 and \(s -1\) is divisible by \(2^3\). So, two possibilities. Therefore, the number of solutions is \BaseError{2}. But in reality, for \(k=24\), we have three solutions: \(s=9, 16, 24\). So, this approach is not capturing all possibilities.
\BaseNote{Assigning no factor to \(s-1\) only requires divisibility by 1, not \(s-1=1\). For \(s=24\), \(s-1=23\) is allowed.}

Alternatively, maybe the number of solutions is equal to the number of divisors of \(k\). For example, \(k=24\) has divisors 1, 2, 3, 4, 6, 8, 12, 24. But this doesn't directly relate.
\par\medskip
Alternatively, perhaps the number of solutions is equal to the number of divisors of \(k\) that are less than or equal to \(k\). But this is vague.
\par\medskip
...
\par\medskip
Let's check \(b=121\) (\(b-1=120\)). 120 has many divisors: 1, 2, 3, 4, 5, 6, 8, 10, 12, 15, 20, 24, 30, 40, 60, 120. So, 16 divisors. Let's see if this leads to more than ten solutions.
\par\medskip
For \(b=121, k=120\). We need to find \(s\) in \([\lceil\sqrt{121}\rceil, 120] = [11, 120]\) such that 120 divides \(s(s -1)\).
\par\medskip
\(120=2^3\cdot 3\cdot 5\). Since \(s\) and \(s -1\) are coprime, the factors of 120 must be split between \(s\) and \(s -1\). Each prime factor must go entirely to one or the other.
\par\medskip
Possible splits:
\par\smallskip
\(s\) divisible by \(8 (2^3), 3\), and 5, and \BaseError{\(s -1=1\)}. But \(s -1=1 \Rightarrow s=2\), which is too small.
\par\smallskip
\(s\) divisible by 8 and 3, and \(s -1\) divisible by 5.
\par\smallskip
\(s\) divisible by 8 and 5, and \(s -1\) divisible by 3.
\par\smallskip
\(s\) divisible by 3 and 5, and \(s -1\) divisible by 8.
\par\smallskip
\(s\) divisible by \(24 (8\cdot 3)\), and \(s -1\) divisible by 5.
\par\smallskip
\(s\) divisible by \(40 (8\cdot 5)\), and \(s -1\) divisible by 3.
\par\smallskip
\(s\) divisible by \(15 (3\cdot 5)\), and \(s -1\) divisible by 8.
\par\smallskip
\(s\) divisible by 120, and \BaseError{\(s -1=1\)}. \BaseError{\(s=121\)}, which is too big.
\BaseNote{Several splits are repeated; the opposite assignments giving \(s=16,25,81\) are missing. Also, \(s-1=1\) gives \(s=2\), not 121.}

Let's solve these congruences.
\par\medskip
Case 1: \(s \equiv 0 \pmod{24}\) and \(s \equiv 1 \pmod{5}\).\par\nopagebreak[4]\smallskip
Let \(s=24m\). Then, \(24m \equiv 1 \pmod{5} \Rightarrow 24m \equiv 1 \pmod{5} \Rightarrow 4m \equiv 1 \pmod{5} \Rightarrow m \equiv 4 \pmod{5}\). So, \(m=5n +4\). Therefore, \(s=24\cdot (5n +4)=120n +96\). Within \(s \leq 120, n=0\): \(s=96\); \(n=1\): \(s=216\) (too big). So, \(s=96\).
\par\medskip
Case 2: \(s \equiv 0 \pmod{40}\) and \(s \equiv 1 \pmod{3}\).\par\nopagebreak[4]\smallskip
Let \(s=40m\). Then, \(40m \equiv 1 \pmod{3} \Rightarrow 1m \equiv 1 \pmod{3} \Rightarrow m \equiv 1 \pmod{3}\). So, \(m=3n +1\). Therefore, \(s=40\cdot (3n +1)=120n +40\). Within \(s \leq 120, n=0\): \(s=40\); \(n=1\): \(s=160\) (too big). So, \(s=40\).
\par\medskip
Case 3: \(s \equiv 0 \pmod{15}\) and \(s \equiv 1 \pmod{8}\).\par\nopagebreak[4]\smallskip
Let \(s=15m\). Then, \(15m \equiv 1 \pmod{8} \Rightarrow 7m \equiv 1 \pmod{8} \Rightarrow m \equiv 7 \pmod{8}\). So, \(m=8n +7\). Therefore, \(s=15\cdot (8n +7)=120n +105\). Within \(s \leq 120, n=0\): \(s=105\); \(n=1\): \(s=225\) (too big). So, \(s=105\).
\par\medskip
Case 4: \(s\) divisible by 8, 3, and 5. \(s=120\), \BaseError{which is too big}.
\par\smallskip
Case 5: \(s=120\), \BaseError{which is too big}.
\BaseNote{The upper endpoint \(s=120\) is included in \([11,120]\).}
Additionally, check if \(s=120\) is allowed. \(s=120, s -1=119\). 120 divides \(120\cdot 119=14280\). \(14280/120=119\), which is integer. So, \(s=120\) is a solution.
\par\medskip
Therefore, solutions are \(s=96, 40, 105, 120\).
\par\medskip
Check if these give valid \(a\) and \(c\).
\par\medskip
For \(s=96\): \(a=(96^2 -96)/120=(9216 -96)/120=9120/120=76\); \(c=96 -76=20\); Check \(a=76\) valid \((b=121, a\leq 120), c=20\) valid. \(n=76\cdot 121 +20=9216\). Check \(76+20=96=\sqrt{9216}\). Yes.
\par\medskip
For \(s=40\): \(a=(1600 -40)/120=1560/120=13\); \(c=40 -13=27\); Check \(a=13\) valid, \(c=27\) valid. \(n=13\cdot 121 +27=\BaseError{1580}\). Check \(13+27=40=\sqrt{1600}\). Yes.
\BaseNote{\(13\cdot121+27=1600\), not 1580.}

For \(s=105\): \(a=(11025 -105)/120=10920/120=91\); \(c=105 -91=14\); Check \(a=91\) valid, \(c=14\) valid. \(n=91\cdot 121 +14=11025\). Check \(91+14=105=\sqrt{11025}\). Yes.
\par\medskip
For \(s=120\): \(a=(14400 -120)/120=14280/120=119\); \(c=120 -119=1\); Check \(a=119\) valid, \(c=1\) valid. \(n=119\cdot 121 +1=14400\). Check \(119+1=120=\sqrt{14400}\). Yes.
\par\medskip

So, \BaseError{count \(=4\) for \(b=121\)}. Still low.
\BaseNote{The correct count is \(2^3-1=7\), not 4.}

This suggests that even with \(b=121\), we only get \BaseError{4 solutions}. Therefore, we need to consider even larger \(b\).
\par\medskip
Alternatively, maybe the number of solutions increases with the number of divisors of \((b -1)\). For example, if \((b -1)=60\), which has 12 divisors, perhaps the count is higher.
\par\medskip
But manually checking each \(b\) is impractical. Given the time I've spent and the lack of progress, I think \BaseError{the answer is likely \(b=121\)}, but according to our previous calculation, it only gives 4 solutions. Therefore, I must have made a mistake in my approach.
\BaseNote{Even the correct count of 7 is below the required 11.}

Wait, perhaps the key is to realize that for each divisor \(d\) of \((b -1)\), there corresponds a solution \BaseError{\(s = d\) and \(s -1 = (b -1)/d\)}, but this requires that \(d -1 = (b -1)/d\), which is a quadratic equation. Solving \(d^2 - d - (b -1)=0\).
\BaseNote{The factors must divide \(s\) and \(s-1\); they need not equal them.}

...
\par\medskip
None of these give integer \(b\). Therefore, \BaseError{there are no solutions where \(s\) and \(s -1\) are divisors of \((b -1)\) in this way.} Therefore, this approach doesn't yield solutions.
\BaseNote{This family is not empty: \(s=d=2\), \(b=3\) gives the valid number \(n=4\).}

Given that I'm stuck, I'll look for patterns in the counts we have:
\par\medskip
...
\par\medskip
\BaseError{\textbf{[Response truncated here; no final answer is provided.]}}
\end{tcolorbox}

\begin{tcolorbox}[
    enhanced,
    breakable,
    colback=violet!3,
    colframe=violet!30,
    colbacktitle=violet!18,
    coltitle=black,
    fonttitle=\bfseries,
    title={Response from D-LoopOPD},
    boxrule=0.8pt,
    arc=2mm,
    outer arc=2mm,
    titlerule=0pt,
    left=8pt,
    right=8pt,
    top=8pt,
    bottom=8pt
]
Okay, so I need to find the smallest integer \(b \geq 2\) such that there are more than ten \(b\)-beautiful numbers. A \(b\)-beautiful number is a two-digit number in base \(b\) where the sum of its digits equals the square root of the number itself. The example given is 81 being 13-beautiful because in base 13, 81 is written as 63 (since \(6 \times 13 + 3 = 81\)), and 6 + 3 = 9, which is \(\sqrt{81}\).
\par\medskip
First, let me make sure I understand the problem correctly. A two-digit number in base \(b\) has the form \(\overline{xy}_b\), which translates to \(x \times b + y\) in decimal. The condition is that the sum of the digits \(x + y\) equals the square root of the number \(n = x \times b + y\). Therefore, the equation we need is:
\par\medskip
\(x + y = \sqrt{n} = \sqrt{x \times b + y}\)
\par\medskip
Since \(n\) must be a perfect square (because its square root is an integer), \(x \times b + y\) must be a perfect square. Also, since \(\overline{xy}_b\) is a two-digit number in base \(b\), \(x\) must be between 1 and \(b-1\), and \(y\) must be between 0 and \(b-1\).
\par\medskip
So, to find all \(b\)-beautiful numbers, we can set up the equation:
\par\medskip
\((x + y)^2 = x \times b + y\)
\par\medskip
Let me rearrange this equation to make it easier to handle. Let's denote \(s = x + y\), so the equation becomes:
\par\medskip
\(s^2 = x \times b + y\)
\par\medskip
But since \(s = x + y\), we can substitute \(y = s - x\) into the equation:
\par\medskip
\(s^2 = x \times b + (s-x),\)
\par\medskip
\(s^2 = x \times (b-1)+s,\)
\par\medskip
\(s^2-s = x \times (b-1),\)
\par\medskip
\(x = \frac{s^2-s}{b-1}.\)
\par\medskip
Since \(x\) must be a positive integer between 1 and \(b - 1\), the right-hand side must also be an integer in that range. Let's note that \(x = \frac{s(s - 1)}{b - 1}\).
\par\medskip
Therefore, for each possible \(s\), we can compute \(x\) and check if it's an integer in the valid range. Also, \(y = s - x\) must be between 0 and \(b - 1\).
\par\medskip
...
\par\medskip
Alternatively, since \(s = x + y\), and \(x \geq 1\), \(y \geq 0\), so \(s \geq 1\). But given that \(n = s^2\), which is a two-digit number in base \(b\), so \(n \geq b\) (since the smallest two-digit number in base \(b\) is \(10_b = b\)). Therefore, \(s^2 \geq b \implies s \geq \sqrt{b}\). Since \(s\) must be an integer, so \(s \geq \lceil \sqrt{b} \rceil\).
\par\medskip
But also, since \(n = x \times b + y \leq (b - 1) \times b + (b - 1) = b^2 - 1\). Therefore, \(s^2 \leq b^2 - 1 \implies s \leq \lfloor \sqrt{b^2 - 1} \rfloor = b - 1\).
\par\medskip
Therefore, \(s\) can range from \(\lceil \sqrt{b} \rceil\) up to \(b - 1\).
\par\medskip
...
\par\medskip
Let me consider that for each \(b\), the number of \(b\)-beautiful numbers is equal to the number of integers \(s\) such that:
\par\medskip
1. \(s\) is an integer with \(\sqrt b\le s\le b-1\).
\par\medskip
2. \(b-1\) divides \(s(s-1)\).
\par\medskip
3. \(x=\frac{s(s-1)}{b-1}\) is an integer between 1 and \(b-1\).
\par\medskip
4. \(y=s-x\) is between 0 and \(b-1\).
\par\medskip
But actually, conditions 1 and 2 and 3 imply condition 4?
\par\medskip
Wait, let's see:
\par\medskip
Since \(x \geq 1\), then \(y = s - x \leq s - 1\). Since \(s \leq b - 1\), \(y \leq (b - 1) - 1 = b - 2 \leq b - 1\). So that's okay.
\par\medskip
Alternatively, since \(y=s-x\ge0\) is equivalent to \(x\le s\), let's check that. We have
\par\medskip
\(x=\frac{s(s-1)}{b-1}\le s \quad\Longleftrightarrow\quad s-1\le b-1 \quad\Longleftrightarrow\quad s\le b.\)
\par\medskip
Which is true since \(s\le b-1\).
\par\medskip
Therefore, as long as \(s\le b-1\) and \(x\) is between 1 and \(b-1\), \(y\) will be between 0 and \(b-1\). Therefore, conditions 1--3 are sufficient.
\par\medskip
In fact, if \(2\le s\le b-1\) and \(b-1\) divides \(s(s-1)\), then \(x\) is a positive integer and \(x<s\). So \(1\le x\le s-1\) and \(1\le y=s-x\le s-1\), which are valid digits. Also,
\par\medskip
\(xb+y=x(b-1)+s=s(s-1)+s=s^2.\)
\par\medskip
Therefore, each such \(s\) gives exactly one \(b\)-beautiful number, and every \(b\)-beautiful number gives one such \(s\). For \(b=2\), there are no such values of \(s\).
\par\medskip
Alternatively, perhaps we can find a formula for the number of such \(s\).
\par\medskip
Note that \(s(s-1)\equiv0\pmod{b-1}\). Since \(s\) and \(s-1\) are consecutive integers, they are coprime. Therefore, each prime power in the factorization of \(b-1\) must divide either \(s\) or \(s-1\).
\par\medskip
...
\par\medskip
Let me denote \(d=b-1\). So we need \(s(s-1)\equiv0\pmod{d}\).
\par\medskip
Therefore, for example, if \(d=12=3\times4\), we can have 3 dividing \(s\) and 4 dividing \(s-1\), or vice versa. In the case of \(s=9\), \(d=12\): 3 divides 9 and 4 divides 8 (since \(s-1=8\)). Therefore, \(3\times4=12\) divides \(9\times8=72\).
\par\medskip
Therefore, in general, for \(d\) to divide \(s(s-1)\), we can write \(d\) as a product of two coprime integers, say \(m\) and \(n\), such that \(m\) divides \(s\) and \(n\) divides \(s-1\). Then, \(s\equiv0\pmod{m}\) and \(s\equiv1\pmod{n}\). By the Chinese Remainder Theorem, since \(m\) and \(n\) are coprime, there exists a solution \(s\equiv a\pmod{d}\), where \(a\) is the solution to the system of congruences.
\par\medskip
Let me denote \(s\equiv k\pmod{d}\), using \(k\) in \([1,d]\). Each residue has exactly one representative \(k\) in this range.
\par\medskip
Therefore, the equation becomes \(k(k-1)\equiv0\pmod{d}\).
\par\medskip
Therefore, for each divisor \(m\) of \(d\), set \(n=d/m\), and check if \(m\) and \(n\) are coprime. Then, the solutions satisfy \(k\equiv0\pmod{m}\) and \(k\equiv1\pmod{n}\). By Chinese Remainder Theorem, since \(m\) and \(n\) are coprime, there is a unique solution modulo \(d\).
\par\medskip
Therefore, the number of solutions \(k\) in \([1,d]\) is equal to the number of such factorizations of \(d\) into ordered coprime pairs \((m,n)\), which is equal to \(2^{\omega(d)}\), where \(\omega(d)\) is the number of distinct prime factors of \(d\). More precisely, for each prime power in the factorization of \(d\), we can assign it to either \(m\) or \(n\). Therefore, if \(d\) has the prime factorization \(d=p_1^{e_1}p_2^{e_2}\cdots p_r^{e_r}\), then the number of ways to split the prime powers into \(m\) and \(n\) (coprime) is \(2^r\), since for each prime power, we can assign it to \(m\) or to \(n\). Therefore, the number of coprime factorizations is \(2^r\), where \(r\) is the number of distinct prime factors of \(d\).
\par\medskip
But wait, in the example with \(d=12\) (\(b=13\)), which has prime factors 2 and 3, so \(r=2\), so \(2^2=4\) solutions. Let's check:
\par\medskip
Let's list all \(k\) in \([1,12]\) such that \(k(k-1)\equiv0\pmod{12}\).
\par\medskip
\(k=1\): \(1\times0\equiv0\pmod{12}\) \(\rightarrow\) yes; \(k=2\): \(2\times1\equiv2\pmod{12}\) \(\rightarrow\) no; \(k=3\): \(3\times2\equiv6\pmod{12}\) \(\rightarrow\) no; \(k=4\): \(4\times3=12\equiv0\pmod{12}\) \(\rightarrow\) yes.
\par\medskip
\(k=5\): \(5\times4=20\equiv8\pmod{12}\) \(\rightarrow\) no; \(k=6\): \(6\times5=30\equiv6\pmod{12}\) \(\rightarrow\) no; \(k=7\): \(7\times6=42\equiv6\pmod{12}\) \(\rightarrow\) no; \(k=8\): \(8\times7=56\equiv8\pmod{12}\) \(\rightarrow\) no.
\par\medskip
\(k=9\): \(9\times8=72\equiv0\pmod{12}\) \(\rightarrow\) yes; \(k=10\): \(10\times9=90\equiv6\pmod{12}\) \(\rightarrow\) no; \(k=11\): \(11\times10=110\equiv2\pmod{12}\) \(\rightarrow\) no; \(k=12\): \(12\times11=132\equiv0\pmod{12}\) \(\rightarrow\) yes.
\par\medskip
Therefore, the solutions are \(k=1,4,9,12\). So 4 solutions, which matches \(2^r\) where \(r=2\) (since 12 has two prime factors). Therefore, the formula holds.
\par\medskip
Therefore, the total number of solutions in \([1,d]\) is \(2^r\), but we need to subtract those \(k<\lceil\sqrt{d+1}\rceil\). For \(2\le k<\lceil\sqrt{d+1}\rceil\), we have \(0<k(k-1)<d\), so none is a solution. Only \(k=1\) is excluded. Therefore, the count we need is \(2^r-1\).
\par\medskip
...
\par\medskip
Given that the number of solutions in \([1,d]\) is \(2^r\), where \(r\) is the number of distinct prime factors of \(d\), to get more than 10 valid solutions, we need \(2^r-1>10\), so \(r\ge4\) (since \(2^4-1=15\)). Therefore, we need \(d\) with at least 4 distinct prime factors.
\par\medskip
Therefore, the minimal \(d\) with 4 distinct prime factors is \(2\times3\times5\times7=210\). Therefore, \(b=211\) is the smallest candidate. Any smaller \(d\) has at most 3 distinct prime factors and therefore at most \(2^3-1=7\) valid solutions.
\par\medskip
...
\par\medskip
Therefore, the formula holds even for prime powers. Therefore, the number of solutions in \([1,d]\) is indeed \(2^r\), where \(r\) is the number of distinct prime factors of \(d\).
\par\medskip
Therefore, for \(d=60=2^2\times3\times5\), \(r=3\), so 8 solutions.
\par\medskip
Wait, let's properly list all solutions for \(d=60\).
\par\medskip
We need \(k(k-1)\equiv0\pmod{60}\).
\par\medskip
This requires that 60 divides \(k(k-1)\). Since \(60=2^2\times3\times5\), and \(k\) and \(k-1\) are coprime, we need to distribute the prime powers between \(k\) and \(k-1\).
\par\medskip
Therefore, total possibilities: for the prime factors, \(2^2\) can be assigned to \(k\) or \(k-1\); 3 can be assigned to \(k\) or \(k-1\); 5 can be assigned to \(k\) or \(k-1\). Therefore, total possibilities: \(2\times2\times2=8\). Each corresponds to a different combination.
\par\medskip
However, solving each case:
\par\medskip
Case 1: \(2^2\rightarrow k,\ 3\rightarrow k-1,\ 5\rightarrow k-1\). The solution is \(k=16\).
\par\medskip
Case 2: \(2^2\rightarrow k,\ 3\rightarrow k,\ 5\rightarrow k-1\). We have \(k\equiv0\pmod{4}\), \(k\equiv0\pmod{3}\), \(k\equiv1\pmod{5}\). Solve \(k\equiv0\pmod{12}\) (since 4 and 3), and \(k\equiv1\pmod{5}\). Let \(k=12m\). Then \(12m\equiv1\pmod{5}\rightarrow 2m\equiv1\pmod{5}\rightarrow m\equiv3\pmod{5}\). So \(m=5n+3\). Thus, \(k=12(5n+3)=60n+36\). In \([1,60]\), \(k=36\).
\par\medskip
Case 3: \(2^2\rightarrow k,\ 3\rightarrow k-1,\ 5\rightarrow k\). We have \(k\equiv0\pmod{4}\), \(k\equiv1\pmod{3}\), \(k\equiv0\pmod{5}\). So \(k\equiv0\pmod{20}\) (since 4 and 5), and \(k\equiv1\pmod{3}\). Let \(k=20m\). Then \(20m\equiv1\pmod{3}\rightarrow 2m\equiv1\pmod{3}\rightarrow m\equiv2\pmod{3}\). So \(m=3n+2\). Thus, \(k=20(3n+2)=60n+40\). In \([1,60]\), \(k=40\).
\par\medskip
Case 4: \(2^2\rightarrow k-1,\ 3\rightarrow k,\ 5\rightarrow k\). We have \(k-1\equiv0\pmod{4}\rightarrow k\equiv1\pmod{4}\), \(k\equiv0\pmod{3}\), \(k\equiv0\pmod{5}\). So \(k\equiv0\pmod{15}\), and \(k\equiv1\pmod{4}\). Let \(k=15m\). Then \(15m\equiv1\pmod{4}\rightarrow 3m\equiv1\pmod{4}\rightarrow m\equiv3\pmod{4}\). So \(m=4n+3\). Thus, \(k=15(4n+3)=60n+45\). In \([1,60]\), \(k=45\).
\par\medskip
Case 5: \(2^2\rightarrow k-1,\ 3\rightarrow k-1,\ 5\rightarrow k\). We have \(k\equiv1\pmod{4}\), \(k\equiv1\pmod{3}\), \(k\equiv0\pmod{5}\). So \(k\equiv1\pmod{12}\) (since 4 and 3), and \(k\equiv0\pmod{5}\). Let \(k=12m+1\). Then \(12m+1\equiv0\pmod{5}\rightarrow 12m\equiv-1\pmod{5}\rightarrow 2m\equiv4\pmod{5}\rightarrow m\equiv2\pmod{5}\). So \(m=5n+2\). Thus, \(k=12(5n+2)+1=60n+25\). In \([1,60]\), \(k=25\).
\par\medskip
Case 6: \(2^2\rightarrow k-1,\ 3\rightarrow k,\ 5\rightarrow k-1\). We have \(k\equiv1\pmod{4}\), \(k\equiv0\pmod{3}\), \(k\equiv1\pmod{5}\). Solve \(k\equiv1\pmod{4}\) and \(k\equiv1\pmod{5}\rightarrow k\equiv1\pmod{20}\). Also, \(k\equiv0\pmod{3}\). So \(k=20m+1\equiv0\pmod{3}\rightarrow 20m\equiv-1\pmod{3}\rightarrow 2m\equiv2\pmod{3}\rightarrow m\equiv1\pmod{3}\). Thus, \(m=3n+1\). So \(k=20(3n+1)+1=60n+21\). In \([1,60]\), \(k=21\).
\par\medskip
Case 7: \(2^2\rightarrow k-1,\ 3\rightarrow k-1,\ 5\rightarrow k-1\). We have \(k-1\equiv0\pmod{4}\), \(k-1\equiv0\pmod{3}\), \(k-1\equiv0\pmod{5}\rightarrow k-1\equiv0\pmod{60}\rightarrow k\equiv1\pmod{60}\). In \([1,60]\), \(k=1\).
\par\medskip
Case 8: \(2^2\rightarrow k,\ 3\rightarrow k,\ 5\rightarrow k\). We have \(k\equiv0\pmod{4}\), \(k\equiv0\pmod{3}\), \(k\equiv0\pmod{5}\rightarrow k\equiv0\pmod{60}\). In \([1,60]\), \(k=60\).
\par\medskip
Therefore, the solutions are \(k=1,16,21,25,36,40,45,60\). That's 8 solutions, which matches the formula.
\par\medskip
So in \([1,60]\), these are the solutions. Now, for \(b=61\) (\(d=60\)), we need \(k\ge\lceil\sqrt{61}\rceil=8\).
\par\medskip
Therefore, the solutions in \([8,60]\) are: \(16,21,25,36,40,45,60\). That's 7 solutions.
\par\medskip
Therefore, for \(b=61\), there are 7 beautiful numbers.
\par\medskip
But we need more than 10. So need to go higher.
\par\medskip
Next, take \(d=105=3\times5\times7\). Then \(r=3\), so \(2^3=8\) solutions. Number of solutions in \([\lceil\sqrt{106}\rceil,105]=[11,105]\). The solutions \(k\) in \([1,105]\) are 8. Need to count how many are \(\ge11\). Only \(k=1\) is below 11, so 7 solutions. Not enough.
\par\medskip
Next, \(d=165=3\times5\times11\). \(r=3\), so 8 solutions. We have \(\lceil\sqrt{166}\rceil=13\). Number of solutions in \([13,165]\): 7, after excluding \(k=1\). Still not enough.
\par\medskip
...
\par\medskip
For \(d=210\), \(\lceil\sqrt{211}\rceil=15\). The number of solutions in \([15,210]\) is equal to the number of solutions in \([1,210]\) minus those in \([1,14]\).
\par\medskip
But to confirm, let's compute the number of solutions in \([1,14]\) for \(d=210\).
\par\medskip
Solutions of \(k(k-1)\equiv0\pmod{210}\). Possible \(k\) from 1 to 14:
\par\medskip
Check \(k=1\): \(1\times0\equiv0\pmod{210}\) \(\rightarrow\) yes; \(k=2\): \(2\times1\equiv2\pmod{210}\) \(\rightarrow\) no; \(k=3\): \(3\times2\equiv6\pmod{210}\) \(\rightarrow\) no; \(k=4\): \(4\times3\equiv12\pmod{210}\) \(\rightarrow\) no.
\par\medskip
\(k=5\): \(5\times4\equiv20\pmod{210}\) \(\rightarrow\) no; \(k=6\): \(6\times5\equiv30\pmod{210}\) \(\rightarrow\) no; \(k=7\): \(7\times6\equiv42\pmod{210}\) \(\rightarrow\) no; \(k=8\): \(8\times7\equiv56\pmod{210}\) \(\rightarrow\) no.
\par\medskip
\(k=9\): \(9\times8\equiv72\pmod{210}\) \(\rightarrow\) no; \(k=10\): \(10\times9\equiv90\pmod{210}\) \(\rightarrow\) no; \(k=11\): \(11\times10\equiv110\pmod{210}\) \(\rightarrow\) no; \(k=12\): \(12\times11\equiv132\pmod{210}\) \(\rightarrow\) no.
\par\medskip
\(k=13\): \(13\times12\equiv156\pmod{210}\) \(\rightarrow\) no; \(k=14\): \(14\times13\equiv182\pmod{210}\) \(\rightarrow\) no.
\par\medskip
Only \(k=1\) is a solution in \([1,14]\).
\par\medskip
Therefore, the number of solutions in \([15,210]\) is \(16-1=15\).
\par\medskip
Therefore, for \(b=211\), there are 15 beautiful numbers, which is more than 10.
\par\medskip
...
\par\medskip
Thus, the least integer \(b \geq 2\) for which there are more than ten \(b\)-beautiful integers is \(\boxed{211}\).
\end{tcolorbox}

\end{document}